\documentclass[11pt]{article}

\usepackage[utf8]{inputenc}
\usepackage[T1]{fontenc}
\usepackage{times}
\usepackage{inconsolata}
\usepackage{microtype}

\usepackage{amsmath}
\usepackage{amssymb}
\usepackage{amsthm}
\usepackage{mathrsfs}

\usepackage{graphicx}
\usepackage{svg}
\usepackage{booktabs}

\usepackage{lipsum}
\usepackage{cite}
\usepackage{url}
\usepackage{pifont}
\usepackage{tikz}
\usepackage{paralist}
\usepackage{multirow}
\usepackage{xifthen}
\usepackage{enumerate}
\usepackage{titlesec}
\usepackage{color}
\usepackage{bm}
\usepackage{caption}
\usepackage{subcaption}
\usepackage{soul}
\usepackage{xspace}
\usepackage{authblk}

\usepackage{algorithm}
\usepackage{algpseudocode}
\usepackage{adjustbox}
\usepackage{colortbl}
\usepackage{dblfloatfix}
\usepackage{multicol}
\usepackage[most]{tcolorbox}
\usepackage{xparse}
\usepackage{hyperref}
\usepackage{cleveref}
\usepackage{capt-of}
\usepackage{etoolbox}
\usepackage{listings}
\usepackage{shadowtext}

\pdfoutput=1

\usepackage{acl}

\makeatletter

\newcounter{reviewer}
\newcounter{point}[reviewer]
\renewcommand{\thereviewer}{\Alph{reviewer}}

\renewcommand{\thepoint}{Comment\,\thereviewer.\arabic{point}}

\newcommand*{\rom}[1]{\MakeUppercase{\romannumeral #1}}
\makeatother

\newcommand{\pptmacro}[1]{%
\textsc{PPT#1}%
\xspace }
\DeclareRobustCommand{\pptagent}{\pptmacro{Agent}}
\DeclareRobustCommand{\ppteval}{\pptmacro{Eval}}

\newcommand{\redtext}[1]{\textcolor[RGB]{176, 36, 24}{#1}}

\newcommand{\greentext}[1]{\textcolor[RGB]{101, 165, 66}{#1}}

\newcommand{\yellowtext}[1]{\textcolor[RGB]{236, 184, 27}{#1}}
\definecolor{darkred}{rgb}{0.5, 0, 0}
\newcommand{\majorrevise}[1]{#1}

\title{\pptagent: Generating and Evaluating Presentations\\ Beyond Text-to-Slides}

\author{ \textbf{Hao Zheng}${}^{1,2,}$
\thanks{~These authors contributed equally}
, \textbf{Xinyan Guan}${}^{1,2,*}$, \textbf{Hao Kong}${}^{3}$, \textbf{Jia Zheng}${}^{1}$, \textbf{Weixiang Zhou}${}^{1}$\\ \textbf{Hongyu Lin${}^{1}$} ,\textbf{Yaojie Lu${}^{1}$},

\textbf{Ben He${}^{1,2}$}, \textbf{Xianpei Han${}^{1}$}, \textbf{Le Sun${}^{1}$}
\\ ${}^{1}$Chinese Information Processing Laboratory, Institute of Software, Chinese
Academy of Sciences\\

${}^{2}$University of Chinese Academy of Sciences \\ ${}^{3}$Shanghai Jiexin
Technology \\ \texttt{\{zhenghao2022,guanxinyan2022,zhengjia, weixiang,hongyu,luyaojie\}@iscas.ac.cn}\\
\texttt{\{xianpei,sunle\}@iscas.ac.cn} ~ \texttt{haokong@knowuheart.com} \texttt{benhe@ucas.edu.cn}
}
\begin{document}
  \maketitle
  \begin{abstract}
    Automatically generating presentations from documents is a challenging task that
    requires accommodating content quality, visual appeal, and structural coherence.
    Existing methods primarily focus on improving and evaluating the content
    quality in isolation, overlooking visual appeal and structural coherence,
    which limits their practical applicability. To address these limitations, we
    propose \pptagent, which comprehensively improves presentation generation through
    a two-stage, edit-based approach inspired by human workflows.
    \pptagent first analyzes reference presentations to extract slide-level functional
    types and content schemas, then drafts an outline and iteratively generates
    editing actions based on selected reference slides to create new slides.
    To comprehensively evaluate the quality of generated presentations, we
    further introduce \ppteval, an evaluation framework that assesses presentations
    across three dimensions: \redtext{Content}, \greentext{Design}, and
    \yellowtext{Coherence}. Results demonstrate that \pptagent significantly
    outperforms existing automatic presentation generation methods across all
    three dimensions. The code
and data are available at \url{https://github.com/icip-cas/PPTAgent}.
  \end{abstract}
\setcounter{page}{1}

  \section{Introduction}

Presentations are a widely used medium for information delivery, valued for their
visual effectiveness in engaging and communicating with audiences. 
However, creating high-quality presentations requires a captivating storyline, well-designed layouts, and rich, compelling content~\citep{Fu_Wang_McDuff_Song_2022}.
Consequently, creating well-rounded presentations requires advanced presentation
skills and significant effort. 
Given the inherent complexity of the presentation creation, there is growing interest in automating the presentation generation process~\citep{mondal2024presentations, maheshwari2024presentations, ge2025autopresent} by leveraging the generalization capabilities of Large Language Models (LLMs) and Multimodal Large Language Models (MLLMs).

\begin{figure}[t]
  \centering
  \includegraphics[width=1.0\linewidth]{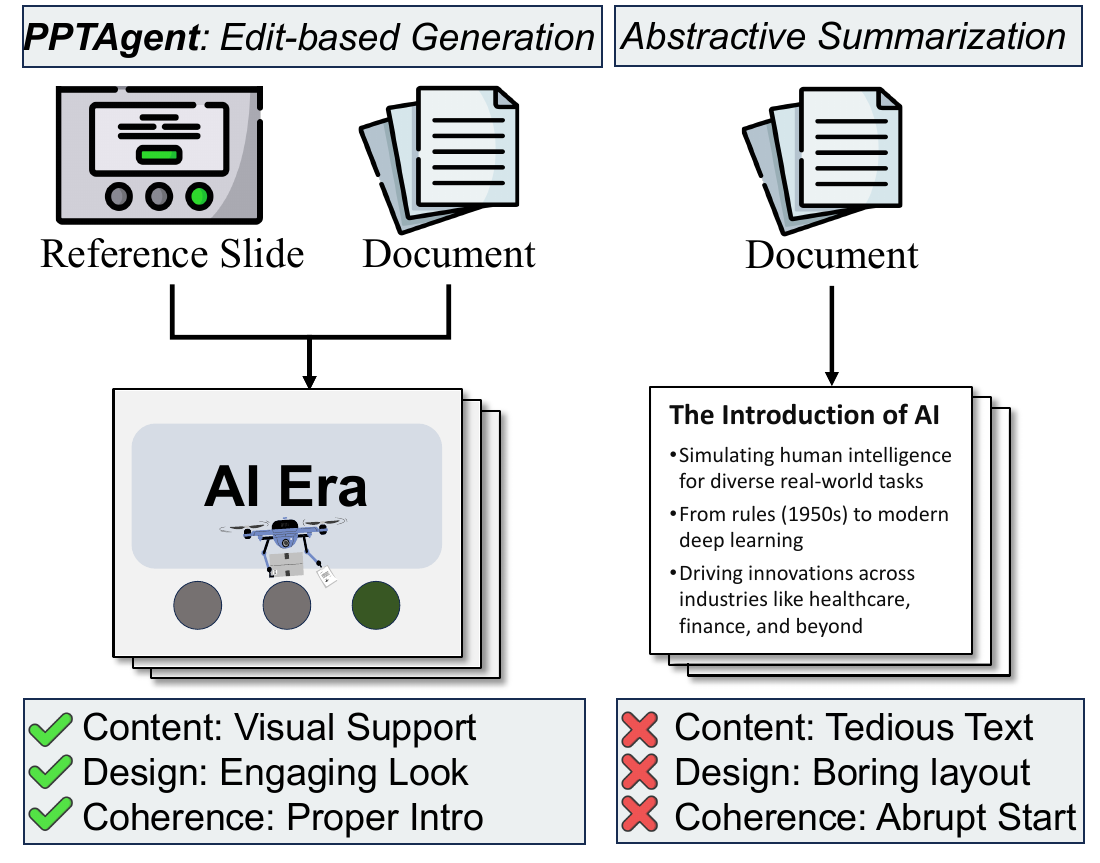}
  \caption{Comparison between our \pptagent approach (left) and the conventional
  abstractive summarization method (right).}
  \label{fig:1}
  \vspace{-10pt}
\end{figure}

\begin{figure*}[t]
  \centering
  \includegraphics[width=\linewidth]{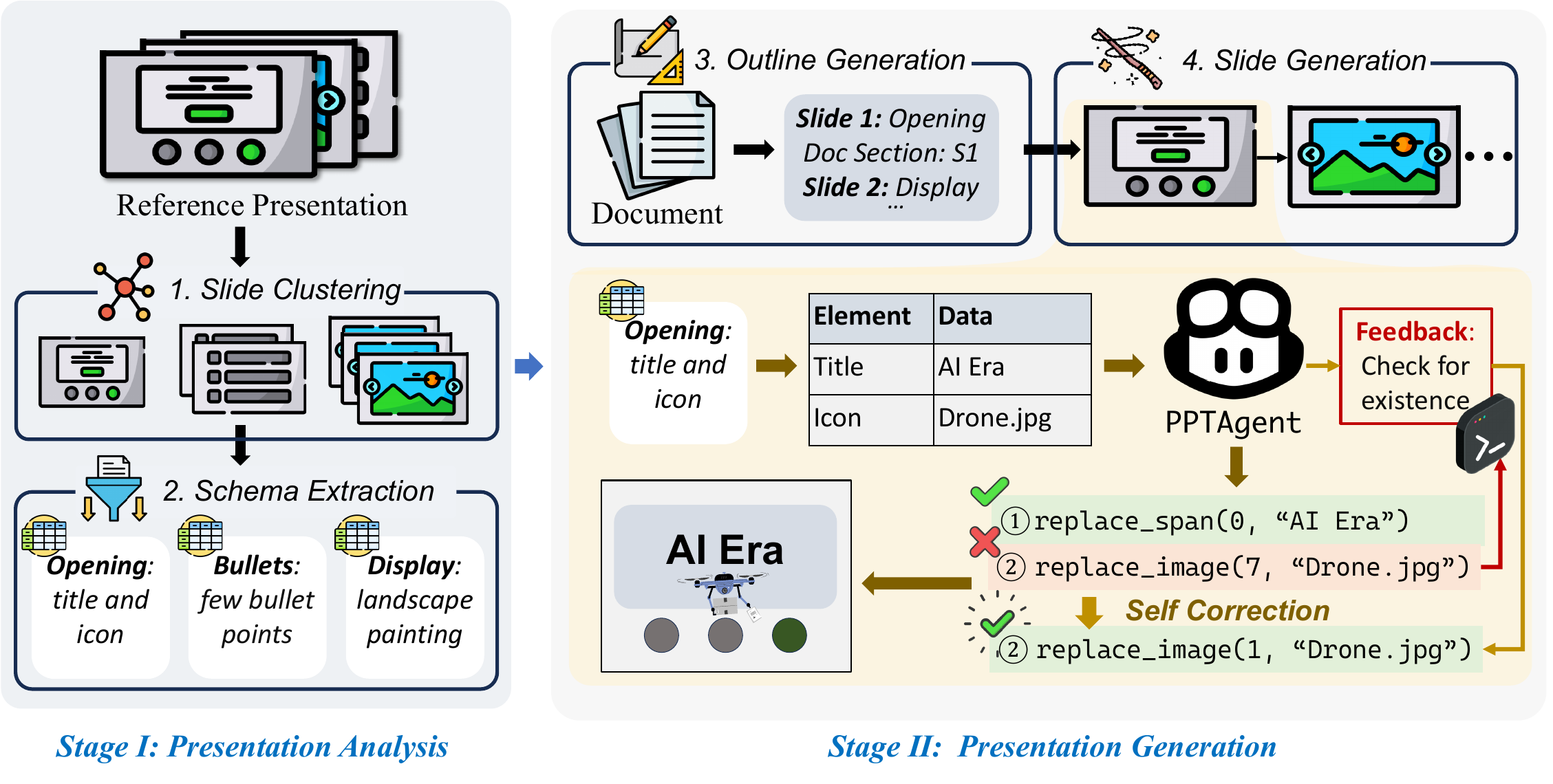}
  \caption{Overview of the \pptagent workflow. \textcolor[rgb]{0, 0.439, 0.753}{\textbf{\textit{Stage\rom{1}: Presentation Analysis}}} involves analyzing the input presentation to
  cluster slides into groups and extract their content schemas. \textcolor[rgb]{0,
  0.439, 0.753}{\textbf{\textit{Stage \rom{2}: Presentation Generation}}}
  generates new presentations guided by the outline, incorporating self-correction mechanisms
  to ensure robustness.}
  \vspace{-5pt}
  \label{fig:method}
\end{figure*}

Existing approaches typically follow a text-to-slides paradigm, which converts
LLM outputs into slides using predefined rules or templates. 
As shown in Figure~\ref{fig:1}, prior studies \citep{mondal2024presentations,sefid2021slidegen} tend to
treat presentation generation as an abstractive summarization task, focusing primarily
on textual content while neglecting the visual-centric nature \citep{Fu_Wang_McDuff_Song_2022} of presentation.
This results in text-heavy and monotonous presentations that fail to engage audiences
effectively \citep{barrick2018image}.

\majorrevise{
Rather than creating complex presentations from scratch in a single pass, human workflows typically involve selecting exemplary slides as references and then summarizing and transferring key content onto them \citep{duarte2010resonate}.
Inspired by this process, we propose \pptagent, which decomposes slide
generation into two phases: selecting the reference slide and editing it step by
step.
However, achieving such an edit-based approach for presentation generation is challenging.
First, due to the layout and modal complexity of presentations, it is difficult for LLMs to directly determine which slides should be referenced.
The key challenge lies in enhancing LLMs' understanding of reference presentations' structure and content patterns.
Second, most presentations are saved in PowerPoint's XML format, as demonstrated in Figure~\ref{fig:xml}, which is
inherently verbose and redundant \citep{gryk2022human}, making it challenging for LLMs to robustly perform editing operations.}

\majorrevise{
To address these challenges, \pptagent operates in two stages.
Stage \rom{1} performs a comprehensive analysis of reference presentations to extract functional types and content schemas of slides, facilitating subsequent reference selection and slide generation.
Stage \rom{2} introduces a suite of edit APIs with HTML-rendered representation that simplifies slide modifications through code interaction \citep{wang2024executable}.
Furthermore, we implement a self-correction mechanism \citep{kamoi2024can} that allows LLMs to iteratively refine generated editing actions based on intermediate results and execution feedback, ensuring robust generation.
As shown in Figure~\ref{fig:method}, we first analyze and cluster reference slides into categories (e.g., opening slides, bullet-point slides).
For each new slide, \pptagent selects an appropriate reference slide (e.g., opening slide for the first slide) and generates a series of editing actions (e.g., replace\_span) to modify it.
}

Due to the lack of a comprehensive evaluation framework, we
propose \ppteval, which adopts the MLLM-as-a-judge paradigm~\cite{chen2024mllm} to evaluate presentations across three dimensions: \redtext{Content}, \greentext{Design}, and \yellowtext{Coherence}\cite{duarte2010resonate}. 
Human evaluations validate the reliability and effectiveness of \ppteval.
Results demonstrate that \pptagent generates high-quality presentations, achieving an average score of 3.67 for the three dimensions in \ppteval.

Our main contributions can be summarized as follows:

  $\bullet$  We propose \pptagent, a framework that redefines automatic presentation generation as an edit-based process guided by reference presentations.

$\bullet$  We introduce \ppteval, a comprehensive evaluation framework that assesses presentations across three dimensions: \redtext{Content}, \greentext{Design}, and \yellowtext{Coherence}.

  $\bullet$  We release\ the \pptagent and \ppteval codebases, along with a new presentation dataset \textit{Zenodo10K}, to support future research.

  \section{\pptagent}
\label{sec:pptagent}
In this section, we formulate the presentation generation task and introduce our
proposed \pptagent framework, which consists of two distinct stages. In stage \rom{1},
we analyze reference presentations through slide clustering and schema extraction,
providing a comprehensive understanding of input presentations that facilitates
subsequent reference selection and slide generation. In stage \rom{2}, we leverage analyzed reference presentations to select reference slides and generate the target presentation for the input document through an iterative editing process. An
overview of our workflow is illustrated in Figure~\ref{fig:method}.

\subsection{Problem Formulation}
\label{sec2-1:formulation}
\pptagent is designed to generate an engaging presentation through an edit-based
process. We provide formal definitions for the conventional method and \pptagent
to highlight their key differences.

The conventional method \citep{bandyopadhyay2024enhancing, mondal2024presentations} for creating each slide $\boldsymbol{S}$ is formalized in
Equation~\ref{eq:convention}. Given the input content $C$, it generates $n$ slide
elements, each defined by its type, content, and styling attributes, such as $(\textrm
{Textbox}, \textrm{"Hello"}, \{\textrm{border}, \textrm{size}, \textrm{position},
\dots\})$.

\begin{equation}
  \label{eq:convention}\boldsymbol{S}= \{e_{1}, e_{2}, \dots, e_{n}\} = f(C)
\end{equation}

\vspace{-0.2cm}
While this conventional method is straightforward, it requires manual
specification of styling attributes, which is challenging for automated generation
\citep{guo2023pptc}. Instead of creating slides from scratch, \pptagent generates a sequence of executable
actions to edit reference slides, thereby preserving their well-designed layouts
and styles. As shown in Equation~\ref{eq:ours}, given the input content $C$ and the
$j$-th reference slide $R_{j}$, which is selected from the reference
presentation, \pptagent generates a sequence of $m$ executable actions, where
each action $a_{i}$ corresponds to a line of executable code.
\begin{equation}
  \label{eq:ours}\boldsymbol{A}= \{a_{1}, a_{2}, \dots, a_{m}\} = g(C, R_{j})
\end{equation}

\subsection{Stage \rom{1}
\label{sec2-2:pre_analysis}: Presentation Analysis}
In this stage, we analyze the reference presentation to guide the reference selection
and slide generation. Firstly, we categorize slides based on their structural and
layout characteristics through slide clustering. Then, we extract content schemas
to identify the content organization of the slide in each cluster, providing a
comprehensive description of slide elements.

\paragraph{Slide Clustering}

Slides can be categorized into two main types based on their functionalities: structural
slides that support the presentation's organization (e.g., opening slides) and content
slides that convey specific information (e.g., bullet-point slides). To
distinguish between these two types, we employ LLMs to segment the presentation accordingly.
For structural slides, we leverage LLMs' long-context capability to analyze all
slides in the input presentation, identifying structural slides, labeling their
structural roles based on their textual features, and grouping them accordingly.
For content slides, we first convert them into images and then apply a hierarchical
clustering approach to group similar slide images. Subsequently, we utilize MLLMs
to analyze the converted slide images, identifying layout patterns within each
cluster. Further details are provided in Appendix~\ref{appendix:layout_analysis}.

\paragraph{Schema Extraction}
After clustering, we further analyzed their content schemas to facilitate the
slide generation. Specifically, we define an extraction framework where each
element is represented by its category, description, and content. This framework
enables a clear and structured representation of each slide. Detailed
instructions are provided in Appendix~\ref{appendix:prompts}, with an example of
the schema shown below.

\begin{table}[h]
\centering
\small

\noindent
\centering 
\resizebox{\linewidth}{!}{
\begin{tabular}{| l | m{0.3\linewidth} | m{0.4\linewidth} |}
    \hline
    \rowcolor{gray!30} \textbf{Category} & \textbf{Description} & \textbf{Data} \\
    \hline
    \textbf{Title} & Main title & Sample Library \\
    \hline
    \textbf{Date} & Date of the event & 15 February 2018 \\
    \hline
    \textbf{Image} & Primary image to illustrate the slide & Picture: Children in a library with \dots \\
    \hline
\end{tabular}}
    
\end{table}

\subsection{Stage \rom{2}
\label{sec2-3:generation}: Presentation Generation}
\pptagent first generates an outline
specifying reference slides and relevant content for each new slide. Then, it
iteratively edits elements from reference slides through edit APIs to create the
target presentation.

\paragraph{Outline Generation}

As shown in Figure~\ref{fig:method}, we utilize LLM to generate a structured
outline consisting of multiple entries. Each entry represents a new slide,
containing the reference slide and relevant document content of the new slide. The
reference slide is selected based on the slide-level functional description in
Stage \rom{1}, while the relevant document content is identified based on the input
document.

\paragraph{Slide Generation}

Guided by the structured outline, slides are generated iteratively based on the
corresponding entries.
For each slide, LLMs incorporate textual content and extracted image captions
from the input document.
The new slide adopts the layout of the reference slide while ensuring consistency
in content and structural clarity.

Specifically, to generate a new slide based on the corresponding entry in the
outline, we design edit-based APIs to enable LLMs to edit the reference slide.
As shown below, these APIs support editing, removing, and duplicating slide elements.
Moreover, given the complexity of the XML format in presentations, which is demonstrated in Appendix \ref{appendix:api}, we render the reference
slide into an HTML representation \citep{feng2024layoutgpt}, offering a more precise
and intuitive format for easier understanding. This HTML-based format, combined with
our edit-based APIs, enables LLMs to perform precise content modifications on
reference slides.

\begin{table}[h]
\centering
\resizebox{\columnwidth}{!}{
\begin{tabular}{lp{0.30\textwidth}}
\hline
\textbf{Function Name} & \textbf{Description} \\ \hline
\texttt{del\_span} & Delete a span. \\ \hline
\texttt{del\_image} & Delete an image element. \\ \hline
\texttt{clone\_paragraph} & Create a duplicate of an existing paragraph. \\ \hline
\texttt{replace\_span} & Replace the content of a span. \\ \hline
\texttt{replace\_image} & Replace the source of image. \\ \hline
\end{tabular}
}

\label{table:api}
\end{table}

\majorrevise{
Furthermore, to enhance robustness during the editing process, we implement a
self-correction mechanism \citep{kamoi2024can}. Specifically, the generated editing
actions are executed within a REPL\footnote{\url{https://en.wikipedia.org/wiki/REPL}}
environment. When actions fail to apply to reference slides, the REPL provides
execution feedback\footnote{\url{https://docs.python.org/3/tutorial/errors.html}}
to assist LLMs in refining their actions. The LLM then analyzes this feedback to
adjust its editing actions \citep{guan2024mitigating, wang2024executable},
enabling iterative refinement until a valid slide is generated or the maximum
retry limit is reached.
}

\section{\ppteval}
\label{sec3:ppteval}

We introduce \ppteval, a comprehensive framework that evaluates presentation quality from multiple dimensions, addressing the absence of reference-free evaluation for presentations. 
The framework provides both numeric scores (1-to-5 scale) and detailed rationales to justify each dimension's assessment.

Grounded in established presentation design principles~\citep{duarte2008slide, duarte2010resonate}, our evaluation framework focuses on three key dimensions, as summarized in Table~\ref{table:criteria}.
Specially, given a generated presentation, we assess the \redtext{content} and \greentext{design} at the slide level, while evaluating \yellowtext{coherence} across the entire presentation.

The complete evaluation process is illustrated in Figure~\ref{fig:ppteval}, with detailed scoring criteria and representative examples provided in Appendix~\ref{appendix:samples_ppteval}.

\begin{figure}[t]
  \centering
  \includegraphics[width=1.0\linewidth]{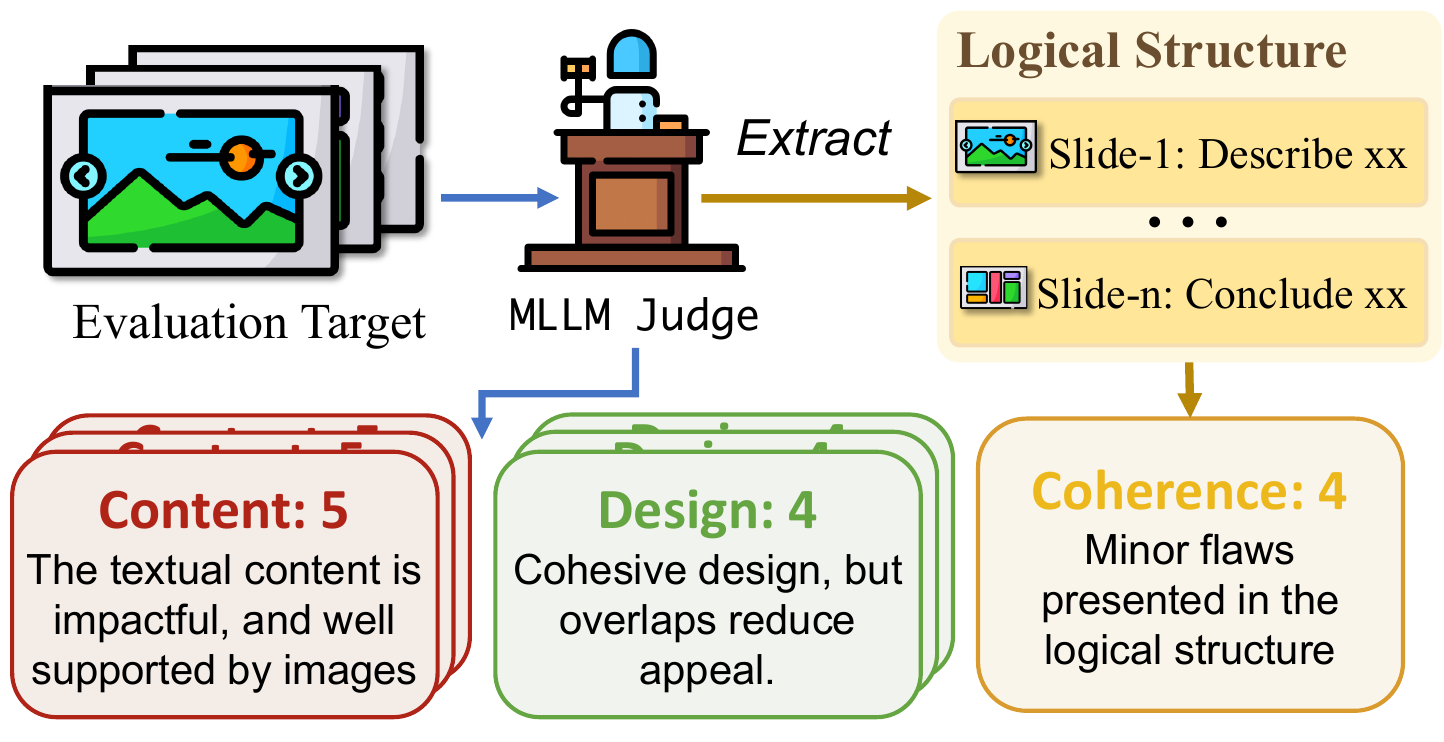}
  \caption{\ppteval assesses presentations from three dimensions: content, design, and coherence.}
  \label{fig:ppteval}
  \vspace{-10pt}
\end{figure}

\begin{table}[h]
\centering
\small
\resizebox{\columnwidth}{!}{
\begin{tabular}{cp{0.35\textwidth}}
\toprule
\textbf{Dimension} & \multicolumn{1}{c}{\textbf{Criteria}} \\
\midrule
\redtext{\textbf{Content}} & Text should be concise and grammatically sound, supported by relevant images.\\
\midrule
\greentext{\textbf{Design}} & Harmonious colors and proper layout ensure readability, while visual elements like geometric shapes enhance the overall appeal.\\
\midrule
\yellowtext{\textbf{Coherence}} & Structure develops progressively, incorporating essential background information.\\
\bottomrule
\end{tabular}
}
\vspace{-2mm}
\caption{The scoring criteria of dimensions in \ppteval, all evaluated in 1-5 scale.}
\vspace{-2mm}
\label{table:criteria}
\end{table}

  \section{Experiment}

\subsection{Dataset}
\label{sec4-1:dataset}

Existing presentation datasets, such as \citet{mondal2024presentations, sefid2021slidegen, sun2021d2s, Fu_Wang_McDuff_Song_2022},
have two main issues. First, they are mostly stored in PDF or JSON formats, which
leads to a loss of semantic information, such as structural relationships and
styling attributes of elements. Additionally, these datasets primarily consist
of academic presentations in artificial intelligence, limiting their diversity.
To address these limitations, we introduce \textit{Zenodo10K}, a new dataset sourced
from Zenodo \citep{https://doi.org/10.25495/7gxk-rd71}, which hosts diverse artifacts
across domains, all under clear licenses. We have curated 10,448 presentations from
this source and made them publicly available to support further research.

Following \citet{mondal2024presentations}, we sample 50 presentations in five
domains to serve as reference presentations. In addition, we collected 50
documents from the same domains to be used as input documents. The sampling criteria and preprocessing details are provided in Appendix \ref{appendix:sample}, while the dataset statistics are summarized in Table~\ref{table:data_stat}.

\begin{table}[t]
  \centering
  \resizebox{\linewidth}{!}{
  \begin{tabular}{lccccc}
    \toprule
    \multirow{2}{*}{\textbf{Domain}} & \multicolumn{2}{c}{\textit{Document}} & \multicolumn{3}{c}{\textit{Presentation}} \\
    \cmidrule(lr){2-3} \cmidrule(lr){4-6}
    & \#Chars & \#Figs & \#Chars & \#Figs & \#Pages \\
    \midrule
    Culture & 12,708 & 2.9 & 6,585 & 12.8 & 14.3 \\
    Education & 12,305 & 5.5 & 3,993 & 12.9 & 13.9 \\
    Science & 16,661 & 4.8 & 5,334 & 24.0 & 18.4 \\
    Society & 13,019 & 7.3 & 3,723 & 9.8 & 12.9 \\
    Tech & 18,315 & 11.4 & 5,325 & 12.9 & 16.8 \\
    \bottomrule
  \end{tabular}}
  \caption{Statistics of the dataset used in our experiments, detailing the number of characters (`\#Chars') and figures (`\#Figs'), as well as the number of pages (`\#Pages').}
  \label{table:data_stat}
\end{table}

\begin{table*}[ht]
    \centering
    
    \resizebox{\linewidth}{!}{
    \begin{tabular}{llcccccccc}
        \toprule
        \multicolumn{2}{c}{\textbf{Configuration}} & \multicolumn{4}{c}{Existing Metrics} & \multicolumn{4}{c}{\ppteval} \\
        \cmidrule(lr){3-6} \cmidrule(lr){7-10}
        Language Model & Vision Model 
        & SR(\%)$\uparrow$ & PPL$\downarrow$ &  ROUGE-L $\uparrow$  &FID$\downarrow$ 
        & \redtext{Content}$\uparrow$ & \greentext{Design}$\uparrow$ & \yellowtext{Coherence}$\uparrow$ & Avg.$\uparrow$ \\
        \midrule
        \rowcolor{gray!20} \multicolumn{10}{c}{\textit{DocPres (rule-based)}} \\
        GPT-4o$_{\texttt{LM}}$ & \multicolumn{1}{c}{--} 
        & -- & 76.42 & 13.28 & -- & 2.98 & 2.33 & 3.24 & 2.85 \\
        Qwen2.5$_{\texttt{LM}}$ & \multicolumn{1}{c}{--} 
        & -- & 100.4 & 13.09 & -- & 2.96 & 2.37 & 3.28 & 2.87 \\
        \midrule
        \rowcolor{gray!20} \multicolumn{10}{c}{\textit{KCTV (template-based)}} \\
        GPT-4o$_{\texttt{LM}}$ & \multicolumn{1}{c}{--} 
        & 80.0 & \underline{68.48} & 10.27 & -- & 2.49 & 2.94 & 3.57 & 3.00 \\
        Qwen2.5$_{\texttt{LM}}$ & \multicolumn{1}{c}{--} 
        & 88.0 & \textbf{41.41} & \textbf{16.76} & -- & 2.55 & 2.95 & 3.36 & 2.95 \\
        \midrule
        \rowcolor{gray!20} \multicolumn{10}{c}{\textit{\pptagent (ours)}} \\
        GPT-4o$_{\texttt{LM}}$ & GPT-4o$_{\texttt{VM}}$ 
        & \textbf{97.8} & 721.54 & 10.17 & 7.48 & \underline{3.25} & 3.24 & \underline{4.39} & \underline{3.62} \\
        Qwen2-VL$_{\texttt{LM}}$ & Qwen2-VL$_{\texttt{VM}}$ 
        & 43.0 & 265.08 & 13.03 & \underline{7.32} & 3.13 & \textbf{3.34} & 4.07 & 3.51 \\
        Qwen2.5$_{\texttt{LM}}$ & Qwen2-VL$_{\texttt{VM}}$ 
        & \underline{95.0} & 496.62 & \underline{14.25} & \textbf{6.20} & \textbf{3.28} & \underline{3.27} & \textbf{4.48} & \textbf{3.67} \\
        \bottomrule
    \end{tabular}}
    \caption{Performance comparison of presentation generation methods, including
   DocPres, KCTV, and our proposed \pptagent. 
    The best/second-best scores are
\textbf{bolded}/\underline{underlined}.
Results are reported using existing metrics, including Success Rate (SR), Perplexity (PPL), Rouge-L, Fr\'{e}chet Inception Distance (FID), and PPTEval.}
    \label{table:mainexp}
\end{table*}

\subsection{Implementation Details}
\label{sec4-2:implementation} \pptagent is implemented with three models: GPT-4o-2024-08-06 (GPT-4o), Qwen2.5-72B-Instruct (Qwen2.5, \citealp{yang2024qwen2}),
and Qwen2-VL-72B-Instruct (Qwen2-VL, \citealp{wang2024qwen2}). These models are categorized
according to the specific modalities they handle, whether textual or visual, as indicated
by their subscripts. Specifically, we define configurations as combinations of a
language model (LM) and a vision model (VM), such as Qwen2.5$_{\texttt{LM}}$+Qwen2-VL$_{\texttt{VM}}$.

Experiment data covers 5 domains, each with 10 input
documents and 10 reference presentations, totaling 500 presentation generation
tasks per configuration (5 domains × 10 input documents × 10 reference
presentations). 
Each slide generation allows a maximum of two self-correction
iterations. We use \citet{chen2024bge} and \citet{wu2020visual} to compute the text
and image embeddings respectively. 
All open-source LLMs are deployed using the
VLLM framework \citep{kwon2023efficient} on NVIDIA A100 GPUs. The
total computational cost for experiments are approximately 500 GPU
hours.

\subsection{Baselines}
\label{sec4-3:baseline} We choose the following baseline methods: \textbf{DocPres}
\citep{bandyopadhyay2024enhancing} propose a rule-based approach that generates narrative-rich
slides through multi-stages, and incorporates images through a similarity-based
mechanism. 
\majorrevise{
\textbf{KCTV} \citep{cachola-etal-2024-knowledge} propose a template-based
method that creates slides in an intermediate format before converting them into
final presentations using predefined templates.
}
The baseline methods operate without vision models since they do not process visual
information. Each configuration generates 50 presentations (5 domains × 10 input
documents), as they do not require reference presentations. Consequently, the FID
metric is excluded from their evaluation.

\subsection{Evaluation Metrics}
\label{sec4-4:metrics} We evaluated the presentation generation using the following
metrics:

$\bullet$ \textbf{Success Rate (SR)} evaluates the robustness of presentation
generation \citep{wu2024gpt}, calculated as the percentage of successfully
completed tasks. For \pptagent, success requires the generation of all slides
without execution errors after self-correction. For KCTV, success is determined
by the successful compilation of the generated LaTeX file. DocPres is excluded from
this evaluation due to its deterministic rule-based conversion.

$\bullet$ \textbf{Perplexity (PPL)} measures the likelihood of the model
generating the given sequence. Using Llama-3-8B \citep{dubey2024llama}, we calculate
the average perplexity across all slides in a presentation. Lower perplexity
scores indicate higher textual fluency \citep{bandyopadhyay2024enhancing}.

\majorrevise{
$\bullet$ \textbf{Rouge-L \citep{lin-2004-rouge}} evaluates textual similarity
by measuring the longest common subsequence between generated and reference
texts. We report the F1 score to balance precision and recall.
}

$\bullet$ \textbf{FID \citep{heusel2017gans}} measures the similarity between
the generated presentation and the reference presentation in the feature space. Due
to the limited sample size, we calculate the FID using a 64-dimensional output
vector.

$\bullet$ \textbf{\ppteval} employs GPT-4o as the judging model to evaluate presentation
quality across three dimensions: content, design, and coherence. We compute content
and design scores by averaging across slides, while coherence is assessed at the
presentation level.

\subsection{Overall Result}
Table~\ref{table:mainexp} presents the performance comparison between \pptagent
and baselines, revealing that:

\begin{table}[t]
   \centering
   
\resizebox{\linewidth}{!}{
\begin{tabular}{lccccc}
       \toprule \textbf{Setting} & SR(\%) & \redtext{Content} & \greentext{Design} & \yellowtext{Coherence} & Avg. \\
       \midrule
       \pptagent & \textbf{95.0} & \textbf{3.28} & 3.27 & \textbf{4.48} & \textbf{3.67} \\
       \textsl{w/o Outline} & 91.0 & 3.24 & \underline{3.30} & 3.36 & 3.30 \\
       \textsl{w/o Schema} & 78.8 & 3.08 & 3.23 & 4.04 & 3.45 \\
       \textsl{w/o Structure} & \underline{92.2} & \textbf{3.28} & 3.25 & 3.45 & 3.32 \\
       \textsl{w/o CodeRender} & 74.6 & \underline{3.27} & \textbf{3.34} & \underline{4.38} & \underline{3.66} \\
    \bottomrule
   \end{tabular}}
      \caption{Ablation analysis of \pptagent utilizing the Qwen2.5$_{\texttt{LM}}$+Qwen2-VL$_{\texttt{VM}}$ configuration, demonstrating the contribution of each components.}
      \label{table:ablation}
      
   \end{table}

\paragraph{\pptagent Significantly Improves Overall Presentation Quality.}
\pptagent demonstrates statistically significant performance improvements over
baseline methods across all three dimensions of \ppteval. Compared to the rule-based
baseline (DocPres), \pptagent exhibits substantial improvements in both the
design and content dimensions (3.34 vs. 2.37, +40.9\%; 3.34 vs. 2.98, +12.1\%),
as presentations generated by the DocPres method show minimal design effort. In comparison
with the template-based baseline (KCTV), \pptagent also achieves notable
improvements in both design and content (3.34 vs. 2.95, +13.2\%; 3.28 vs. 2.55,
+28.6\%), underscoring the efficacy of the edit-based paradigm. Most notably,
\pptagent shows a significant enhancement in the coherence dimension (4.48 vs.
3.57, +25.5\% for DocPres; 4.48 vs. 3.28, +36.6\% for KCTV). This improvement
can be attributed to \pptagent’s comprehensive analysis of the
structural role of slides.

\paragraph{\pptagent Exhibits Robust Generation Performance.}
Our approach empowers LLMs to produce well-rounded presentations with remarkable
success rate, achieving $\geq 95\%$ success rate for both Qwen2.5$_{\texttt{LM}}$+Qwen2-VL$_{\texttt{VM}}$
and GPT-4o$_{\texttt{LM}}$+GPT-4o$_{\texttt{VM}}$, which is a significant
improvement compared to KCTV (97.8\% vs. 88.0\%). Moreover, detailed performance
of Qwen2.5$_{\texttt{LM}}$+Qwen2-VL$_{\texttt{VM}}$ across various domains is
illustrated in Table~\ref{table:domain_detail}, underscoring the versatility and
robustness of our approach.

\paragraph{\ppteval Demonstrates Superior Evaluation Capability.}
Traditional metrics like PPL and ROUGE-L demonstrate inconsistent evaluation trends compared to \ppteval.
For instance, KCTV achieves a high ROUGE-L (16.76) but a low content score (2.55), while our method shows the opposite trend with ROUGE-L (14.25) and content score (3.28).
Moreover, we observe that ROUGE score overemphasizes textual alignment with source documents, potentially compromising the expressiveness of presentations. 
Most importantly, \ppteval advances beyond existing metrics through its dual capability of reference-free design assessment and holistic evaluation of presentation coherence.
Further agreement evaluation is shown in Section~\ref{sec5-5:Effectiveness}.

\section{Analysis}
\subsection{Ablation Study}
We conducted ablation studies across four settings: (1) randomly selecting a
slide as the reference (w/o Outline), (2) omitting structural slides during outline
generation (w/o Structure), (3) replacing the slide representation with the
method proposed by \citet{guo2023pptc} (w/o CodeRender), and (4) removing
guidance from the content schema (w/o Schema). All experiments were conducted using
the Qwen2.5${_\texttt{LM}}$+Qwen2-VL${_\texttt{VM}}$ configuration.

\begin{figure*}[t]
    \centering
    \includegraphics[width=1.0\linewidth]{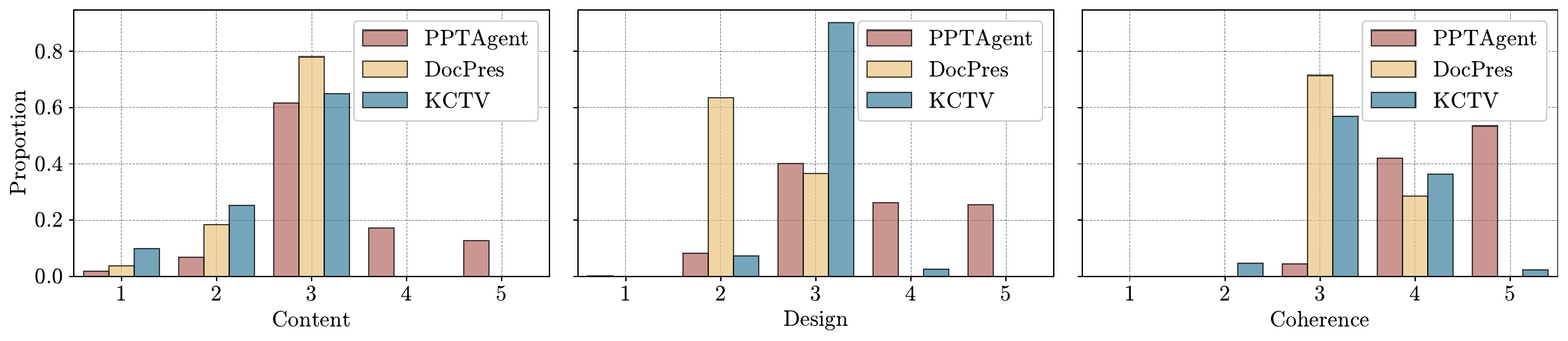}
    \caption{Score distributions of presentations generated by \pptagent, DocPres,
    and KCTV across the three evaluation dimensions: \redtext{Content}, \greentext{Design},
    and \yellowtext{Coherence}, as assessed by \ppteval.}
    \label{fig:quantitative}
\end{figure*}

\begin{figure}[t]
    \centering
    \includegraphics[width=1.0\linewidth]{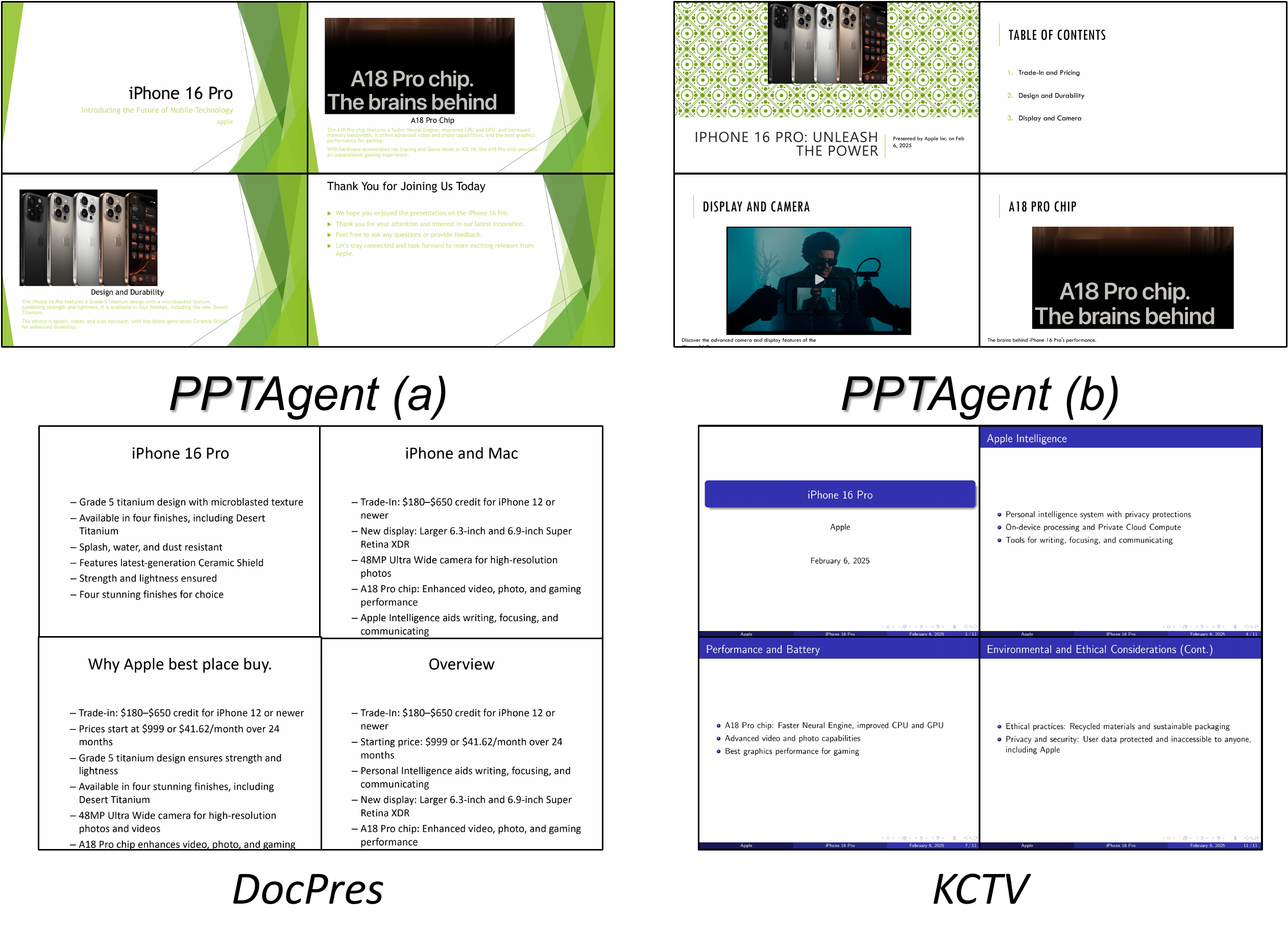}
    \caption{Comparative analysis of presentation generation across different
    methods. \pptagent generates under different reference presentations,
    indicated as \pptagent\textit{ (a)} and \pptagent\textit{ (b)}.}
    \label{fig:qualitative_analysis}
    \vspace{-10pt}
\end{figure}
As demonstrated in Table~\ref{table:ablation}, our experiments reveal two key
findings: 1) \textbf{ The HTML-based representation significantly reduces
interaction complexity}, evidenced by the substantial decrease in success rate from
95.0\% to 74.6\% when removing the Code Render component.
2) \textbf{ The
presentation analysis is crucial for generation quality}, as removing the
outline and structural slides significantly degrades coherence (from 4.48 to 3.36/3.45)
and eliminating the slide schema reduces the success rate from 95.0\% to 78.8\%.

\majorrevise{
\subsection{Case Study}
\label{sec5-2:Qualitative} We present representative examples of presentations generated
under different configurations in Figure~\ref{fig:qualitative_analysis}.
\pptagent demonstrates superior presentation quality across multiple dimensions. First,
it effectively incorporates visual elements with contextually appropriate image
placements, while maintaining concise and well-structured slide content. Second,
it exhibits diversity in generating visually engaging slides under diverse references.
In contrast, baseline methods (DocPres and KCTV) produce predominantly text-based
slides with limited visual variation, constrained by their rule-based or
template-based paradigms.
}
\begin{figure}[t]
    \centering
    \includegraphics[width=\linewidth]{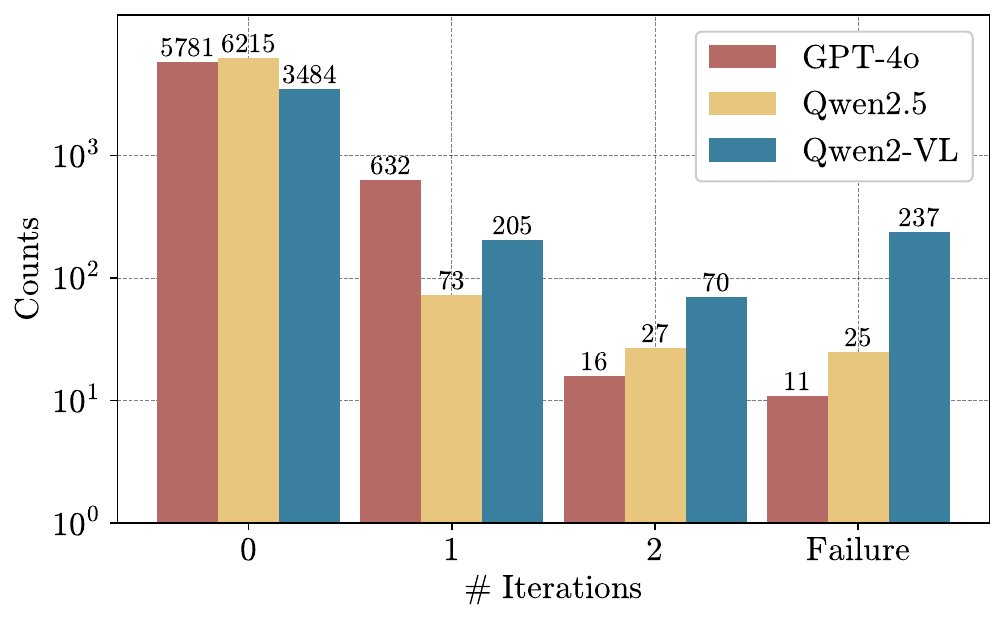}
    \caption{The number of iterative self-corrections required to generate a
    single slide under different models.}
    
    \label{fig:error_analysis}
\end{figure}
\majorrevise{
\subsection{Score Distribution}
\label{sec5-3:Quantitative} We further investigated the score distribution of generated
presentations to compare the performance characteristics across methods, as
shown in Figure~\ref{fig:quantitative}. Constrained by their rule-based or template-based
paradigms, baseline methods exhibit limited diversity in both content and design
dimensions, with scores predominantly concentrated at levels 2 and 3. In contrast,
\pptagent demonstrates a more dispersed score distribution, with the majority of presentations
(>80\%) achieving scores of 3 or higher in these dimensions.
Furthermore, due to \pptagent's comprehensive consideration of structural slides,
it achieves notably superior coherence scores, with over 80\% of the
presentations receiving scores above 4.
}
\subsection{Effectiveness of Self-Correction}
\label{sec5-4:error_analysis} Figure~\ref{fig:error_analysis} illustrates the number
of iterations required to generate a slide using different language models.
Although GPT-4o exhibits superior self-correction capabilities compared to Qwen2.5,
Qwen2.5 encounters fewer errors in the first generation. Additionally, we observed
that Qwen2-VL experiences errors more frequently and has poorer self-correction capabilities,
likely due to its multimodal post-training \citep{wang2024qwen2}. Ultimately, all
three models successfully corrected more than half of the errors, demonstrating
that our iterative self-correction mechanism effectively ensures the success of
the generation process.

\subsection{Agreement Evaluation}
\label{sec5-5:Effectiveness}
\begin{figure}[t]
    \centering
    \includegraphics[width=0.8\linewidth]{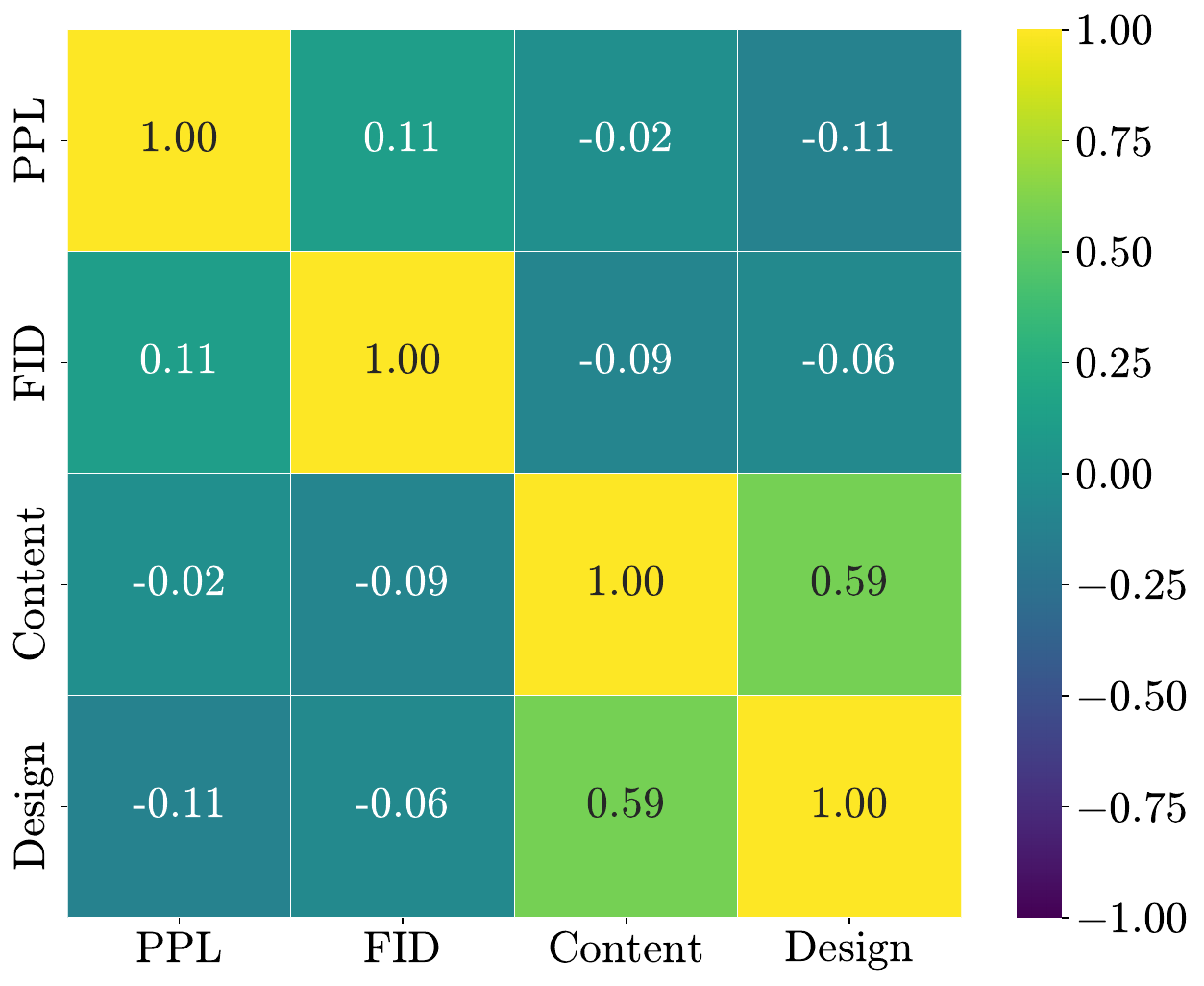}
    \caption{Correlation heatmap between existing automated evaluation metrics
    along with the content and design dimension in \ppteval.}
    \label{fig:correlation_heatmap}
    \vspace{-10pt}
\end{figure}

\paragraph{\ppteval with Human Preferences}

Despite \citet{chen2024mllm} have highlighted the impressive human-like
discernment of LLMs in various generation tasks. However, it remains crucial to
assess the correlation between LLM evaluations and human evaluations in the
context of presentations. This necessity arises from findings by \citet{laskar-etal-2024-systematic},
which indicate that LLMs may not be adequate evaluators for complex tasks. Table~\ref{table:agreement}
shows the correlation of ratings between humans and LLMs. The average Pearson correlation
of 0.71 exceeds the scores of other evaluation methods \citep{kwan2024mteval}, indicating
that \ppteval aligns well with human preferences.

\paragraph{\ppteval with Existing Metrics}
We analyzed the relationships between \ppteval's content and design dimensions and
existing metrics through Pearson correlation analysis, as shown in Figure~\ref{fig:correlation_heatmap}.
The Pearson correlation coefficients reveal that current metrics are ineffective
for presentation evaluation. Specifically, PPL primarily measures text fluency but
performs poorly on slide content due to its inherent fragmented nature,
frequently producing outlier measurements. Similarly, while ROUGE-L and FID
quantify similarity to reference text and presentations respectively, these metrics
inadequately assess content and design quality, as high conformity to references
does not guarantee presentation effectiveness. These weak correlations highlight
the necessity of \ppteval for robust and comprehensive presentation evaluation that
considers both content quality and design effectiveness.
  \section{Related Works}
\paragraph{Automated Presentation Generation}
Recent proposed methods for slide generation can be categorized into rule-based and
template-based based on how they handle element placement and styling. Rule-based
methods, such as those proposed by \citet{mondal2024presentations} and \citet{bandyopadhyay2024enhancing},
often focus on enhancing textual content but neglect the visual-centric nature
of presentations, leading to outputs that lack engagement. Template-based
methods, including \citet{cachola-etal-2024-knowledge} and industrial solutions
like \href{https://tongyi.aliyun.com/aippt}{Tongyi}, rely on predefined templates
to create visually appealing presentations. However, their dependence on extensive
manual effort for template annotation significantly limits scalability and flexibility.

\paragraph{LLM Agent}
Numerous studies \citep{li2024appagent,deng2024mind2web,tang2025worldcoder} have
explored the potential of LLMs to act as agents assisting humans in a wide array
of tasks. For example, \citet{wang2024executable} demonstrate the capability of
LLMs to accomplish tasks by generating executable actions. Furthermore, \citet{guo2023pptc}
demonstrated the potential of LLMs in automating presentation-related tasks through
API integration.

\paragraph{LLM as a Judge}
LLMs have exhibited strong capabilities in instruction following and context perception,
which has led to their widespread adoption as judges \citep{liu-etal-2023-g, zheng2023judging}.
\citet{chen2024mllm} demonstrated the feasibility of using MLLMs as judges,
while \citet{kwan2024mteval} proposed a multi-dimensional evaluation framework.
Additionally, \citet{ge2025autopresent} investigated the use of LLMs for
assessing single-slide quality. However, they did not evaluate presentation quality
from a holistic perspective.
  \begin{table}[t]
\centering
\resizebox{\linewidth}{!}{

\begin{tabular}{lcccc}
\toprule
\textbf{Correlation} & Content & Design & Coherence & Avg. \\
\midrule
\textbf{Pearson} & 0.70 & 0.90 & 0.55 & 0.71 \\
\textbf{Spearman} & 0.73 & 0.88 & 0.57 & 0.74 \\
\bottomrule
\end{tabular}}
\caption{The correlation scores between human ratings and LLM ratings under different dimensions (Coherence, Content, Design). All presented data of similarity exhibit a p-value below 0.05, indicating a statistically significant level of confidence.}
\label{table:agreement}
\end{table}
  \section{Conclusion}
 In this paper, we introduce \pptagent, which conceptualizes presentation generation as a two-stage presentation editing task completed through LLMs' abilities to understand and generate code. Moreover, we propose \ppteval to provide quantitative metrics for assessing presentation quality. Our experiments across data from multiple domains have demonstrated the superiority of our method. This research provides a new paradigm for generating slides under unsupervised conditions and offers insights for future work in presentation generation.

  \section*{Limitations}
  While \pptagent demonstrates promising capabilities in presentation generation,
  several limitations remain. First, despite achieving a high success rate (>95\%)
  on our dataset, the model occasionally fails to generate presentations, which
  could limit its reliability. Second, although we can provide high-quality
  preprocessed presentations as references, the quality of generated presentations
  is still influenced by the input reference presentation, which may lead to
  suboptimal outputs. Third, although \pptagent shows improvements in layout
  optimization compared to prior approaches, it does not fully utilize visual information
  to refine the slide design. This manifests in occasional design flaws, such as
  overlapping elements, which can compromise the readability of generated slides.
  Future work should focus on enhancing the robustness, reducing reference
  dependency, and better incorporating visual information into the generation process.

  \section*{Ethical Considerations}
  In the construction of \textit{Zenodo10K}, we utilized the publicly available API
  to scrape data while strictly adhering to the licensing terms associated with each
  artifact. Specifically, artifacts that were not permitted for modification or commercial
  use under their respective licenses were filtered out to ensure compliance with
  intellectual property rights. Additionally, all annotation personnel involved in
  the project were compensated at rates exceeding the minimum wage in their respective
  cities, reflecting our commitment to fair labor practices and ethical
  standards.

  \bibliography{custom.bib}

\begin{thebibliography}{36}
\providecommand{\natexlab}[1]{#1}

\bibitem[{Bandyopadhyay et~al.(2024)Bandyopadhyay, Maheshwari, Natarajan, and Saxena}]{bandyopadhyay2024enhancing}
Sambaran Bandyopadhyay, Himanshu Maheshwari, Anandhavelu Natarajan, and Apoorv Saxena. 2024.
\newblock Enhancing presentation slide generation by llms with a multi-staged end-to-end approach.
\newblock \emph{arXiv preprint arXiv:2406.06556}.

\bibitem[{Barrick et~al.(2018)Barrick, Davis, and Winkler}]{barrick2018image}
Andrea Barrick, Dana Davis, and Dana Winkler. 2018.
\newblock Image versus text in powerpoint lectures: Who does it benefit?
\newblock \emph{Journal of Baccalaureate Social Work}, 23(1):91--109.

\bibitem[{Cachola et~al.(2024)Cachola, Cucerzan, Herring, Mijovic, Oveson, and Jauhar}]{cachola-etal-2024-knowledge}
Isabel~Alyssa Cachola, Silviu Cucerzan, Allen Herring, Vuksan Mijovic, Erik Oveson, and Sujay~Kumar Jauhar. 2024.
\newblock \href {https://aclanthology.org/2024.findings-emnlp.906} {Knowledge-centric templatic views of documents}.
\newblock In \emph{Findings of the Association for Computational Linguistics: EMNLP 2024}, pages 15460--15476, Miami, Florida, USA. Association for Computational Linguistics.

\bibitem[{Chen et~al.(2024{\natexlab{a}})Chen, Chen, Zhang, Liu, Wang, Zhou, Zhang, Zhou, Wan, and Sun}]{chen2024mllm}
Dongping Chen, Ruoxi Chen, Shilin Zhang, Yinuo Liu, Yaochen Wang, Huichi Zhou, Qihui Zhang, Pan Zhou, Yao Wan, and Lichao Sun. 2024{\natexlab{a}}.
\newblock Mllm-as-a-judge: Assessing multimodal llm-as-a-judge with vision-language benchmark.
\newblock \emph{arXiv preprint arXiv:2402.04788}.

\bibitem[{Chen et~al.(2024{\natexlab{b}})Chen, Xiao, Zhang, Luo, Lian, and Liu}]{chen2024bge}
Jianlv Chen, Shitao Xiao, Peitian Zhang, Kun Luo, Defu Lian, and Zheng Liu. 2024{\natexlab{b}}.
\newblock Bge m3-embedding: Multi-lingual, multi-functionality, multi-granularity text embeddings through self-knowledge distillation.
\newblock \emph{arXiv preprint arXiv:2402.03216}.

\bibitem[{Deng et~al.(2024)Deng, Gu, Zheng, Chen, Stevens, Wang, Sun, and Su}]{deng2024mind2web}
Xiang Deng, Yu~Gu, Boyuan Zheng, Shijie Chen, Sam Stevens, Boshi Wang, Huan Sun, and Yu~Su. 2024.
\newblock Mind2web: Towards a generalist agent for the web.
\newblock \emph{Advances in Neural Information Processing Systems}, 36.

\bibitem[{Duarte(2008)}]{duarte2008slide}
Nancy Duarte. 2008.
\newblock \emph{Slide: ology: The art and science of creating great presentations}, volume~1.
\newblock O'Reilly Media Sebastapol.

\bibitem[{Duarte(2010)}]{duarte2010resonate}
Nancy Duarte. 2010.
\newblock \emph{Resonate: Present visual stories that transform audiences}.
\newblock John Wiley \& Sons.

\bibitem[{Dubey et~al.(2024)Dubey, Jauhri, Pandey, Kadian, Al-Dahle, Letman, Mathur, Schelten, Yang, Fan et~al.}]{dubey2024llama}
Abhimanyu Dubey, Abhinav Jauhri, Abhinav Pandey, Abhishek Kadian, Ahmad Al-Dahle, Aiesha Letman, Akhil Mathur, Alan Schelten, Amy Yang, Angela Fan, et~al. 2024.
\newblock The llama 3 herd of models.
\newblock \emph{arXiv preprint arXiv:2407.21783}.

\bibitem[{{European Organization For Nuclear Research} and {OpenAIRE}(2013)}]{https://doi.org/10.25495/7gxk-rd71}
{European Organization For Nuclear Research} and {OpenAIRE}. 2013.
\newblock \href {https://doi.org/10.25495/7GXK-RD71} {Zenodo}.

\bibitem[{Feng et~al.(2024)Feng, Zhu, Fu, Jampani, Akula, He, Basu, Wang, and Wang}]{feng2024layoutgpt}
Weixi Feng, Wanrong Zhu, Tsu-jui Fu, Varun Jampani, Arjun Akula, Xuehai He, Sugato Basu, Xin~Eric Wang, and William~Yang Wang. 2024.
\newblock Layoutgpt: Compositional visual planning and generation with large language models.
\newblock \emph{Advances in Neural Information Processing Systems}, 36.

\bibitem[{Fu et~al.(2022)Fu, Wang, McDuff, and Song}]{Fu_Wang_McDuff_Song_2022}
Tsu-Jui Fu, William~Yang Wang, Daniel McDuff, and Yale Song. 2022.
\newblock \href {https://doi.org/10.1609/aaai.v36i1.19943} {Doc2ppt: Automatic presentation slides generation from scientific documents}.
\newblock \emph{Proceedings of the AAAI Conference on Artificial Intelligence}, 36(1):634--642.

\bibitem[{Ge et~al.(2025)Ge, Wang, Zhou, Peng, Subramanian, Tan, Sap, Suhr, Fried, Neubig et~al.}]{ge2025autopresent}
Jiaxin Ge, Zora~Zhiruo Wang, Xuhui Zhou, Yi-Hao Peng, Sanjay Subramanian, Qinyue Tan, Maarten Sap, Alane Suhr, Daniel Fried, Graham Neubig, et~al. 2025.
\newblock Autopresent: Designing structured visuals from scratch.
\newblock \emph{arXiv preprint arXiv:2501.00912}.

\bibitem[{Gryk(2022)}]{gryk2022human}
Michael~Robert Gryk. 2022.
\newblock Human readability of data files.
\newblock \emph{Balisage series on markup technologies}, 27.

\bibitem[{Guan et~al.(2024)Guan, Liu, Lin, Lu, He, Han, and Sun}]{guan2024mitigating}
Xinyan Guan, Yanjiang Liu, Hongyu Lin, Yaojie Lu, Ben He, Xianpei Han, and Le~Sun. 2024.
\newblock Mitigating large language model hallucinations via autonomous knowledge graph-based retrofitting.
\newblock In \emph{Proceedings of the AAAI Conference on Artificial Intelligence}, volume~38, pages 18126--18134.

\bibitem[{Guo et~al.(2023)Guo, Zhang, Liang, Zhao, and Nan}]{guo2023pptc}
Yiduo Guo, Zekai Zhang, Yaobo Liang, Dongyan Zhao, and Duan Nan. 2023.
\newblock Pptc benchmark: Evaluating large language models for powerpoint task completion.
\newblock \emph{arXiv preprint arXiv:2311.01767}.

\bibitem[{Heusel et~al.(2017)Heusel, Ramsauer, Unterthiner, Nessler, and Hochreiter}]{heusel2017gans}
Martin Heusel, Hubert Ramsauer, Thomas Unterthiner, Bernhard Nessler, and Sepp Hochreiter. 2017.
\newblock Gans trained by a two time-scale update rule converge to a local nash equilibrium.
\newblock \emph{Advances in neural information processing systems}, 30.

\bibitem[{Kamoi et~al.(2024)Kamoi, Zhang, Zhang, Han, and Zhang}]{kamoi2024can}
Ryo Kamoi, Yusen Zhang, Nan Zhang, Jiawei Han, and Rui Zhang. 2024.
\newblock When can llms actually correct their own mistakes? a critical survey of self-correction of llms.
\newblock \emph{Transactions of the Association for Computational Linguistics}, 12:1417--1440.

\bibitem[{Kwan et~al.(2024)Kwan, Zeng, Jiang, Wang, Li, Shang, Jiang, Liu, and Wong}]{kwan2024mteval}
Wai-Chung Kwan, Xingshan Zeng, Yuxin Jiang, Yufei Wang, Liangyou Li, Lifeng Shang, Xin Jiang, Qun Liu, and Kam-Fai Wong. 2024.
\newblock \href {https://arxiv.org/abs/2401.16745} {Mt-eval: A multi-turn capabilities evaluation benchmark for large language models}.
\newblock \emph{Preprint}, arXiv:2401.16745.

\bibitem[{Kwon et~al.(2023)Kwon, Li, Zhuang, Sheng, Zheng, Yu, Gonzalez, Zhang, and Stoica}]{kwon2023efficient}
Woosuk Kwon, Zhuohan Li, Siyuan Zhuang, Ying Sheng, Lianmin Zheng, Cody~Hao Yu, Joseph Gonzalez, Hao Zhang, and Ion Stoica. 2023.
\newblock Efficient memory management for large language model serving with pagedattention.
\newblock In \emph{Proceedings of the 29th Symposium on Operating Systems Principles}, pages 611--626.

\bibitem[{Laskar et~al.(2024)Laskar, Alqahtani, Bari, Rahman, Khan, Khan, Jahan, Bhuiyan, Tan, Parvez, Hoque, Joty, and Huang}]{laskar-etal-2024-systematic}
Md~Tahmid~Rahman Laskar, Sawsan Alqahtani, M~Saiful Bari, Mizanur Rahman, Mohammad Abdullah~Matin Khan, Haidar Khan, Israt Jahan, Amran Bhuiyan, Chee~Wei Tan, Md~Rizwan Parvez, Enamul Hoque, Shafiq Joty, and Jimmy Huang. 2024.
\newblock \href {https://doi.org/10.18653/v1/2024.emnlp-main.764} {A systematic survey and critical review on evaluating large language models: Challenges, limitations, and recommendations}.
\newblock In \emph{Proceedings of the 2024 Conference on Empirical Methods in Natural Language Processing}, pages 13785--13816, Miami, Florida, USA. Association for Computational Linguistics.

\bibitem[{Li et~al.(2024)Li, Zhang, Yang, Fu, Cheng, Chen, Chen, and Wei}]{li2024appagent}
Yanda Li, Chi Zhang, Wanqi Yang, Bin Fu, Pei Cheng, Xin Chen, Ling Chen, and Yunchao Wei. 2024.
\newblock Appagent v2: Advanced agent for flexible mobile interactions.
\newblock \emph{arXiv preprint arXiv:2408.11824}.

\bibitem[{Lin(2004)}]{lin-2004-rouge}
Chin-Yew Lin. 2004.
\newblock \href {https://aclanthology.org/W04-1013/} {{ROUGE}: A package for automatic evaluation of summaries}.
\newblock In \emph{Text Summarization Branches Out}, pages 74--81, Barcelona, Spain. Association for Computational Linguistics.

\bibitem[{Liu et~al.(2023)Liu, Iter, Xu, Wang, Xu, and Zhu}]{liu-etal-2023-g}
Yang Liu, Dan Iter, Yichong Xu, Shuohang Wang, Ruochen Xu, and Chenguang Zhu. 2023.
\newblock \href {https://doi.org/10.18653/v1/2023.emnlp-main.153} {{G}-eval: {NLG} evaluation using gpt-4 with better human alignment}.
\newblock In \emph{Proceedings of the 2023 Conference on Empirical Methods in Natural Language Processing}, pages 2511--2522, Singapore. Association for Computational Linguistics.

\bibitem[{Maheshwari et~al.(2024)Maheshwari, Bandyopadhyay, Garimella, and Natarajan}]{maheshwari2024presentations}
Himanshu Maheshwari, Sambaran Bandyopadhyay, Aparna Garimella, and Anandhavelu Natarajan. 2024.
\newblock Presentations are not always linear! gnn meets llm for document-to-presentation transformation with attribution.
\newblock \emph{arXiv preprint arXiv:2405.13095}.

\bibitem[{Mondal et~al.(2024)Mondal, Shwetha, Natarajan, Garimella, Bandyopadhyay, and Boyd-Graber}]{mondal2024presentations}
Ishani Mondal, S~Shwetha, Anandhavelu Natarajan, Aparna Garimella, Sambaran Bandyopadhyay, and Jordan Boyd-Graber. 2024.
\newblock Presentations by the humans and for the humans: Harnessing llms for generating persona-aware slides from documents.
\newblock In \emph{Proceedings of the 18th Conference of the European Chapter of the Association for Computational Linguistics (Volume 1: Long Papers)}, pages 2664--2684.

\bibitem[{Sefid et~al.(2021)Sefid, Mitra, and Giles}]{sefid2021slidegen}
Athar Sefid, Prasenjit Mitra, and Lee Giles. 2021.
\newblock Slidegen: an abstractive section-based slide generator for scholarly documents.
\newblock In \emph{Proceedings of the 21st ACM Symposium on Document Engineering}, pages 1--4.

\bibitem[{Sun et~al.(2021)Sun, Hou, Wang, Zhang, and Wang}]{sun2021d2s}
Edward Sun, Yufang Hou, Dakuo Wang, Yunfeng Zhang, and Nancy~XR Wang. 2021.
\newblock D2s: Document-to-slide generation via query-based text summarization.
\newblock \emph{arXiv preprint arXiv:2105.03664}.

\bibitem[{Tang et~al.(2025)Tang, Key, and Ellis}]{tang2025worldcoder}
Hao Tang, Darren Key, and Kevin Ellis. 2025.
\newblock Worldcoder, a model-based llm agent: Building world models by writing code and interacting with the environment.
\newblock \emph{Advances in Neural Information Processing Systems}, 37:70148--70212.

\bibitem[{VikParuchuri(2023)}]{github}
VikParuchuri. 2023.
\newblock \href {https://github.com/VikParuchuri/marker/} {marker}.

\bibitem[{Wang et~al.(2024{\natexlab{a}})Wang, Bai, Tan, Wang, Fan, Bai, Chen, Liu, Wang, Ge et~al.}]{wang2024qwen2}
Peng Wang, Shuai Bai, Sinan Tan, Shijie Wang, Zhihao Fan, Jinze Bai, Keqin Chen, Xuejing Liu, Jialin Wang, Wenbin Ge, et~al. 2024{\natexlab{a}}.
\newblock Qwen2-vl: Enhancing vision-language model's perception of the world at any resolution.
\newblock \emph{arXiv preprint arXiv:2409.12191}.

\bibitem[{Wang et~al.(2024{\natexlab{b}})Wang, Chen, Yuan, Zhang, Li, Peng, and Ji}]{wang2024executable}
Xingyao Wang, Yangyi Chen, Lifan Yuan, Yizhe Zhang, Yunzhu Li, Hao Peng, and Heng Ji. 2024{\natexlab{b}}.
\newblock Executable code actions elicit better llm agents.
\newblock \emph{arXiv preprint arXiv:2402.01030}.

\bibitem[{Wu et~al.(2020)Wu, Xu, Dai, Wan, Zhang, Yan, Tomizuka, Gonzalez, Keutzer, and Vajda}]{wu2020visual}
Bichen Wu, Chenfeng Xu, Xiaoliang Dai, Alvin Wan, Peizhao Zhang, Zhicheng Yan, Masayoshi Tomizuka, Joseph Gonzalez, Kurt Keutzer, and Peter Vajda. 2020.
\newblock \href {https://arxiv.org/abs/2006.03677} {Visual transformers: Token-based image representation and processing for computer vision}.
\newblock \emph{Preprint}, arXiv:2006.03677.

\bibitem[{Wu et~al.(2024)Wu, Yang, Li, Zhang, Liu, Guibas, Lin, and Wetzstein}]{wu2024gpt}
Tong Wu, Guandao Yang, Zhibing Li, Kai Zhang, Ziwei Liu, Leonidas Guibas, Dahua Lin, and Gordon Wetzstein. 2024.
\newblock Gpt-4v (ision) is a human-aligned evaluator for text-to-3d generation.
\newblock In \emph{Proceedings of the IEEE/CVF Conference on Computer Vision and Pattern Recognition}, pages 22227--22238.

\bibitem[{Yang et~al.(2024)Yang, Yang, Zhang, Hui, Zheng, Yu, Li, Liu, Huang, Wei et~al.}]{yang2024qwen2}
An~Yang, Baosong Yang, Beichen Zhang, Binyuan Hui, Bo~Zheng, Bowen Yu, Chengyuan Li, Dayiheng Liu, Fei Huang, Haoran Wei, et~al. 2024.
\newblock Qwen2. 5 technical report.
\newblock \emph{arXiv preprint arXiv:2412.15115}.

\bibitem[{Zheng et~al.(2023)Zheng, Chiang, Sheng, Zhuang, Wu, Zhuang, Lin, Li, Li, Xing et~al.}]{zheng2023judging}
Lianmin Zheng, Wei-Lin Chiang, Ying Sheng, Siyuan Zhuang, Zhanghao Wu, Yonghao Zhuang, Zi~Lin, Zhuohan Li, Dacheng Li, Eric Xing, et~al. 2023.
\newblock Judging llm-as-a-judge with mt-bench and chatbot arena.
\newblock \emph{Advances in Neural Information Processing Systems}, 36:46595--46623.

\end{thebibliography}

  \clearpage
\appendix

\section{Data Preprocessing}
\label{appendix:sample} To maintain a reasonable cost, we selected presentations
ranging from 12 to 64 pages and documents with text lengths from 2,048 to 20,480
characters. We extracted both textual and visual content from the source documents
using \citet{github}. The extracted text was then organized into sections. For
visual content, we generated image captions to assist in relevant image selection
through textual descriptions. To minimize redundancy, we identified and removed duplicate
images if their image embeddings had a cosine similarity score exceeding 0.85.
For slide-level deduplication, we removed individual slides if their text embeddings
had a cosine similarity score above 0.8 compared to the preceding slide, as
suggested by \citet{Fu_Wang_McDuff_Song_2022}.

\section{Details of \ppteval}
\label{appendix:samples_ppteval} We recruited four graduate students through a Shanghai-based
crowdsourcing platform to evaluate a total of 250 presentations: 50 randomly
selected from \textit{Zenodo10K} representing real-world presentations, along
with two sets of 100 presentations generated by the baseline method and our
approach respectively. Following the evaluation framework proposed by \ppteval,
assessments were conducted across three dimensions using the scoring criteria
detailed in Appendix~\ref{appendix:prompts}. Evaluators were provided with converted
slide images, scored them individually, and then discussed the results to reach a
consensus on the final scores.

\begin{figure}[h]
    \centering
    \includegraphics[width=1.0\linewidth]{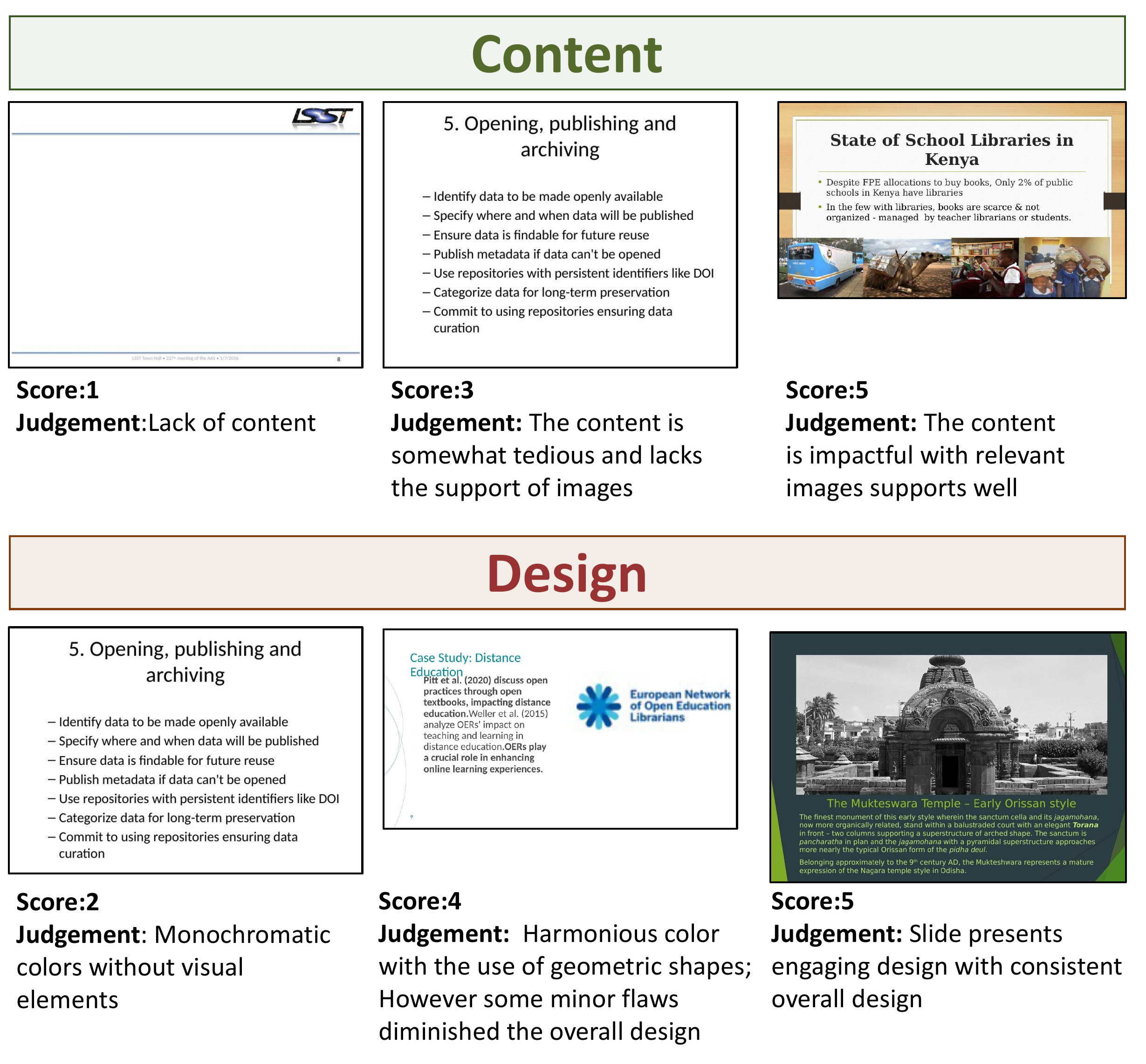}
    \caption{Scoring Examples of \ppteval.}
    \label{fig:example_ppteval}
\end{figure}

Moreover, We measured inter-rater agreement using Fleiss' Kappa, with an average
score of 0.59 across three dimensions (0.61, 0.61, 0.54 for Content, Design, and
Coherence, respectively) indicating satisfactory agreement \citep{kwan2024mteval}
among evaluators. Representative scoring examples are shown in Figure~\ref{fig:example_ppteval}.

We provided detailed illustration as below:

\paragraph{\redtext{Content:}}
The content dimension evaluates the information presented on the slides, focusing
on both text and images. We assess content quality from three perspectives: the amount
of information, the clarity and quality of textual content, and the support provided
by visual content. High-quality textual content is characterized by clear, impactful
text that conveys the proper amount of information. Additionally, images should complement
and reinforce the textual content, making the information more accessible and
engaging. To evaluate content quality, we employ MLLMs on slide images, as slides
cannot be easily comprehended in a plain text format.

\paragraph{\greentext{Design:}}
Good design not only captures attention but also enhances content delivery. We evaluate
the design dimension based on three aspects: color schemes, visual elements, and
overall design. Specifically, the color scheme of the slides should have clear
contrast to highlight the content while maintaining harmony. The use of visual elements,
such as geometric shapes, can make the slide design more expressive. Finally, good
design should adhere to basic design principles, such as avoiding overlapping
elements and ensuring that design does not interfere with content delivery.

\paragraph{\yellowtext{Coherence:}}
Coherence is essential for maintaining audience engagement in a presentation. We
evaluate coherence based on the logical structure and the contextual information
provided. Effective coherence is achieved when the model constructs a captivating
storyline, enriched with contextual information that enables the audience to
follow the content seamlessly. We assess coherence by analyzing the logical structure
and contextual information extracted from the presentation.

\section{Detailed Performance of \pptagent}
We present a detailed performance analysis of Qwen2.5${\texttt{LM}}$+Qwen2-VL${\texttt{VM}}$
across various domains in Table~\ref{table:domain_detail}. Additionally, Table~\ref{table:weighted_ablation} and \ref{table:weighted_mainexp}
show the success rate-weighted performance, where failed generations receive a
\ppteval score of 0, demonstrating that a lower success rate significantly impacts
the overall effectiveness of the method.

As demonstrated in Table~\ref{table:weighted_mainexp}. GPT-4o consistently demonstrates
outstanding performance across various evaluation metrics, highlighting its
advanced capabilities. While Qwen2-VL exhibits limitations in linguistic proficiency
due to the trade-offs from multimodal post-training, GPT-4o maintains a clear
advantage in handling language tasks. However, the introduction of Qwen2.5 successfully
mitigates these linguistic deficiencies, bringing its performance on par with
GPT-4o, and achieving the best performance. This underscores the significant
potential of open-source LLMs as competitive and highly capable presentation
agents.

\section{Slide Clustering}
\label{appendix:layout_analysis} 
\begin{figure}[h]
    \centering
    \includegraphics[width=1.0\linewidth]{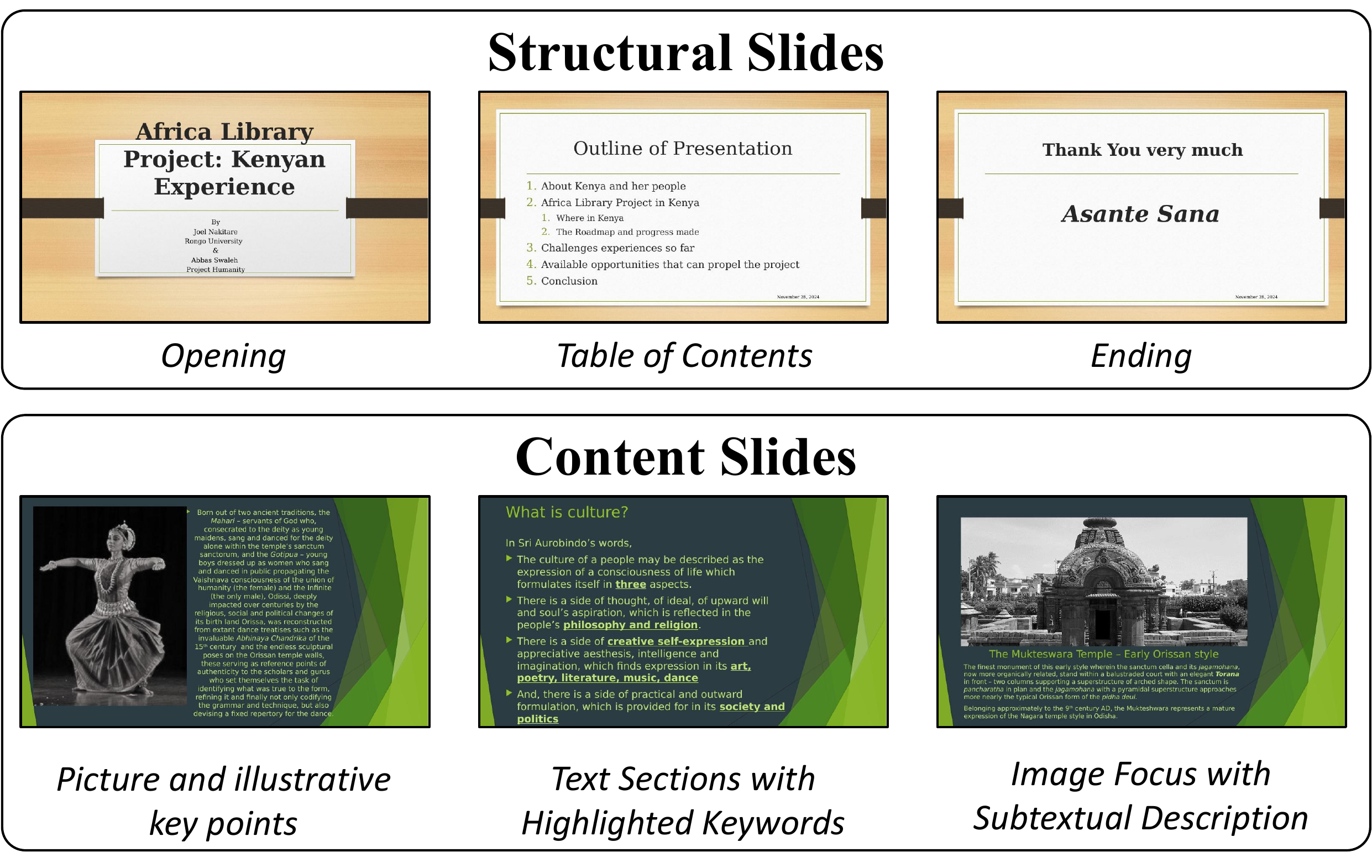}
    \caption{Example of slide clusters.}
    \label{fig:example_layouts}
\end{figure}

We present our hierarchical clustering
algorithm for layout analysis in Algorithm ~\ref{alg:simplified_clustering},
where slides are grouped into clusters using a similarity threshold $\theta$ of
0.65. To focus exclusively on layout patterns and minimize interference from specific
content, we preprocess the slides by replacing text content with a placeholder
character (``a'') and substituting image elements with solid-color backgrounds. Then,
we compute the similarity matrix using cosine similarity based on the ViT
embeddings of converted slide images between each slide pair. Figure~\ref{fig:example_layouts}
illustrates representative examples from the resulting slide clusters.

\section{Code Interaction}
\label{appendix:api}

\begin{figure}[h]
    \centering
    \includegraphics[width=1.0\linewidth]{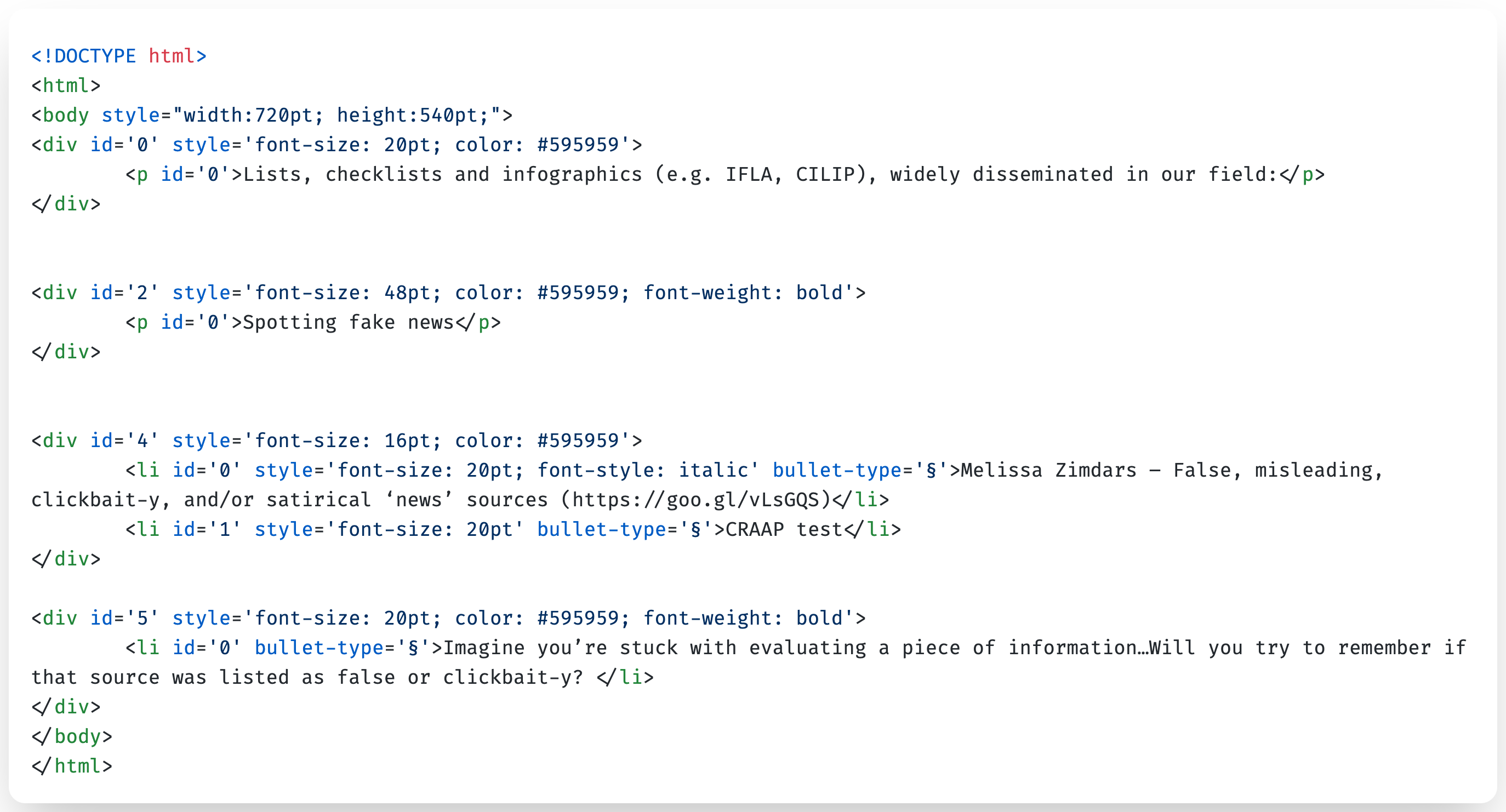}
    \caption{Example of rendering a slide into HTML format.}
    \label{fig:html}
\end{figure}

\begin{figure}[h]
    \centering
    \includegraphics[width=1.0\linewidth]{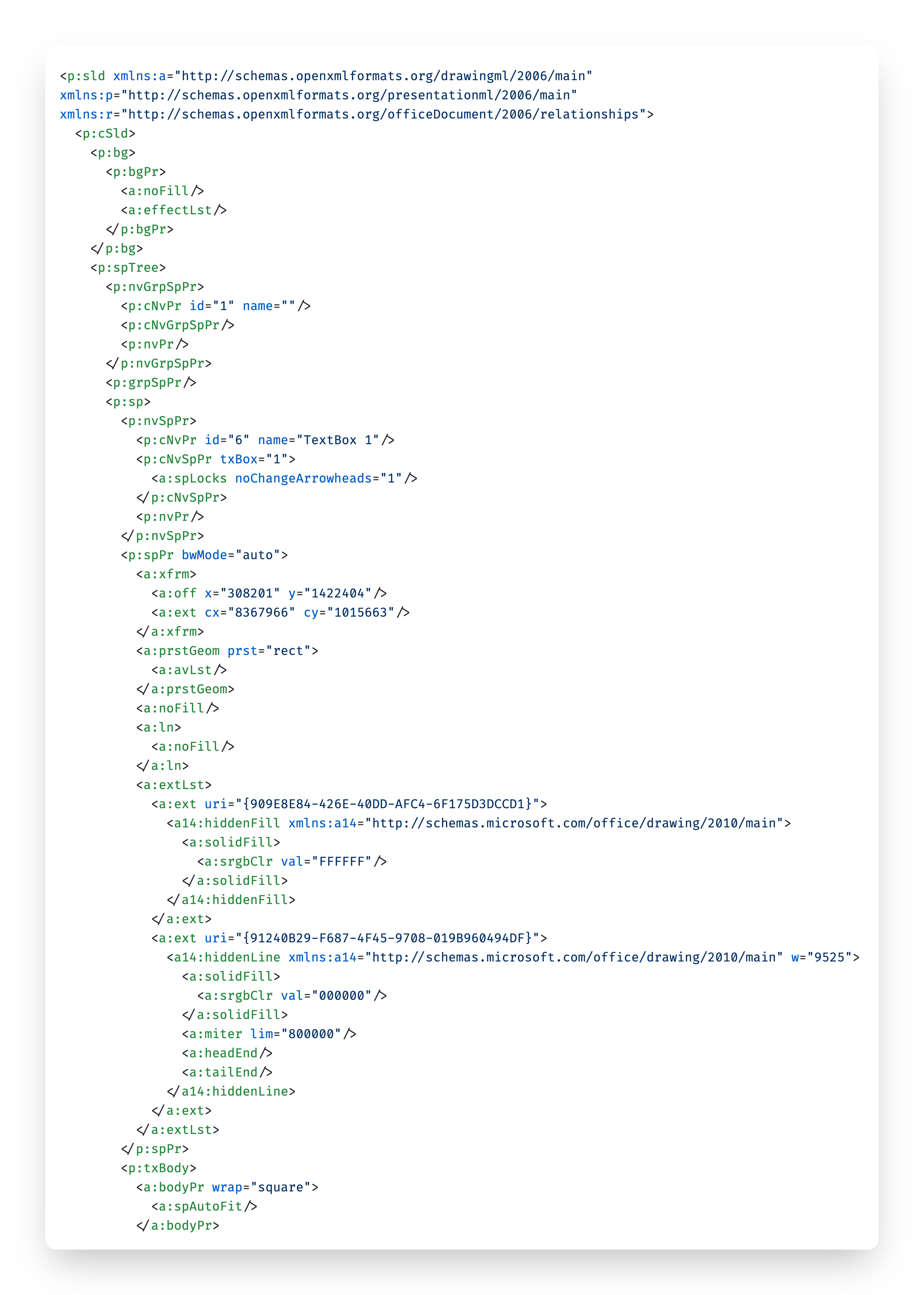}
    \caption{The first 60 lines of the XML representation of a presentation
    slide (out of 1,006 lines).}
    \label{fig:xml}
\end{figure}

For visual reference, Figure~\ref{fig:html} illustrates a slide rendered in HTML
format, while Figure~\ref{fig:xml} displays its excerpt (first 60 lines) of the XML
representation (out of 1,006 lines).

\section{Prompts}
\label{appendix:prompts}

\subsection{Prompts for Presentation Analysis}
\begin{figure}
    \centering
    \includegraphics[width=1.0\linewidth]{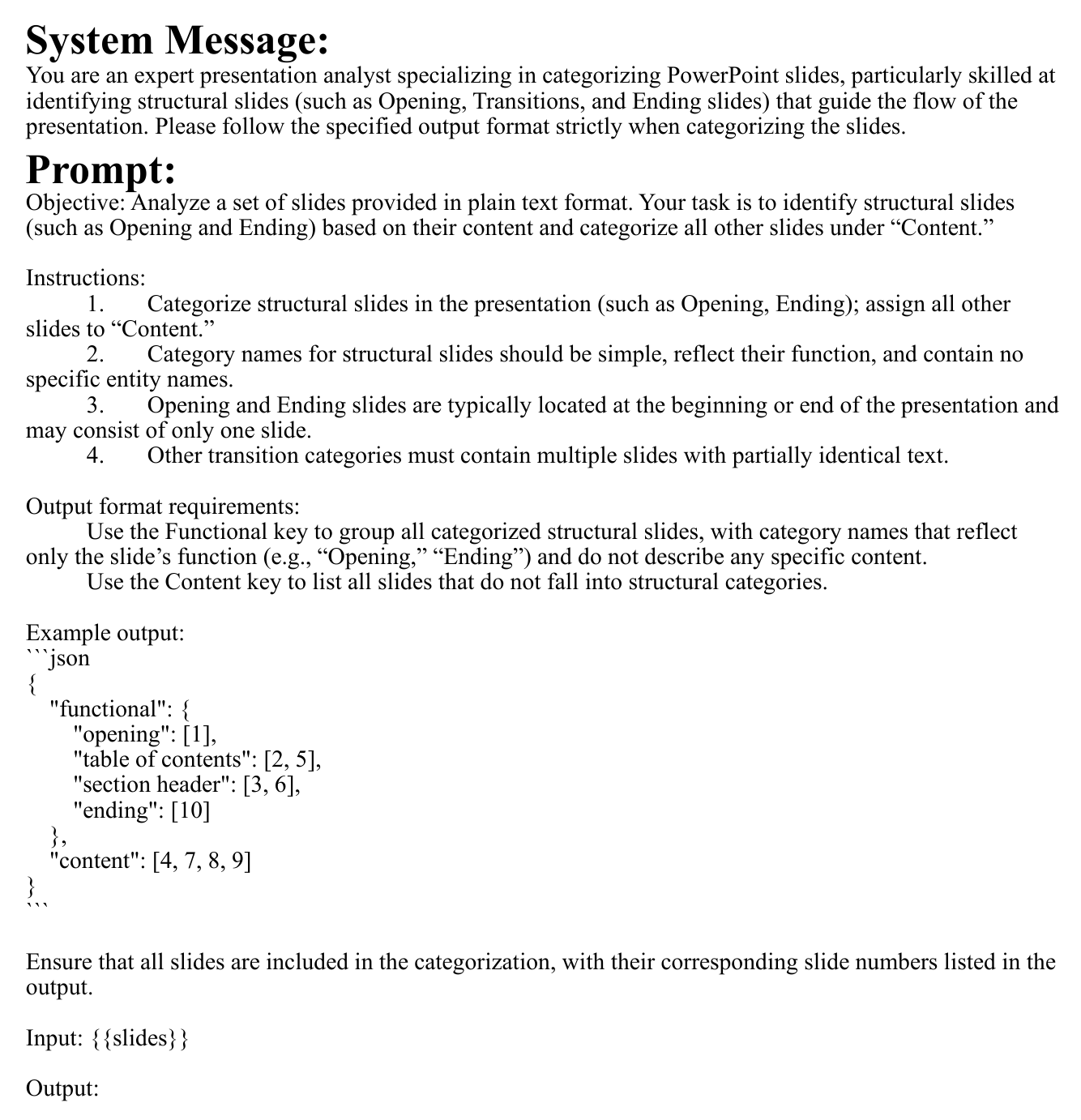}
    \caption{Illustration of the prompt used for clustering structural slides.}
    \label{fig:structure_cluster}
\end{figure}

\begin{figure}
    \centering
    \includegraphics[width=1.0\linewidth]{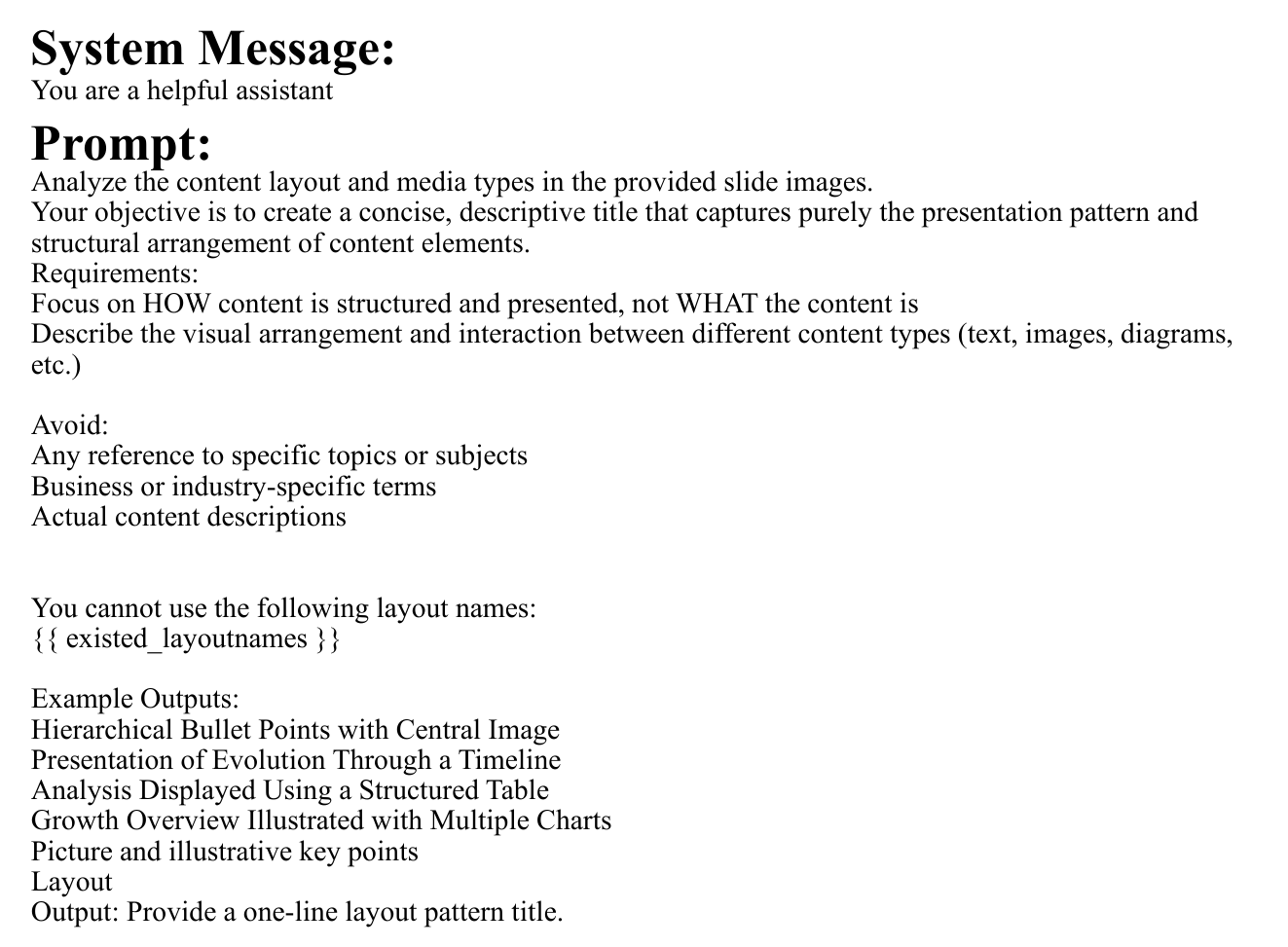}
    \caption{Illustration of the prompt used to infer layout patterns.}
    \label{fig:layout_analysis}
\end{figure}

\begin{figure}
    \centering
    \includegraphics[width=1.0\linewidth]{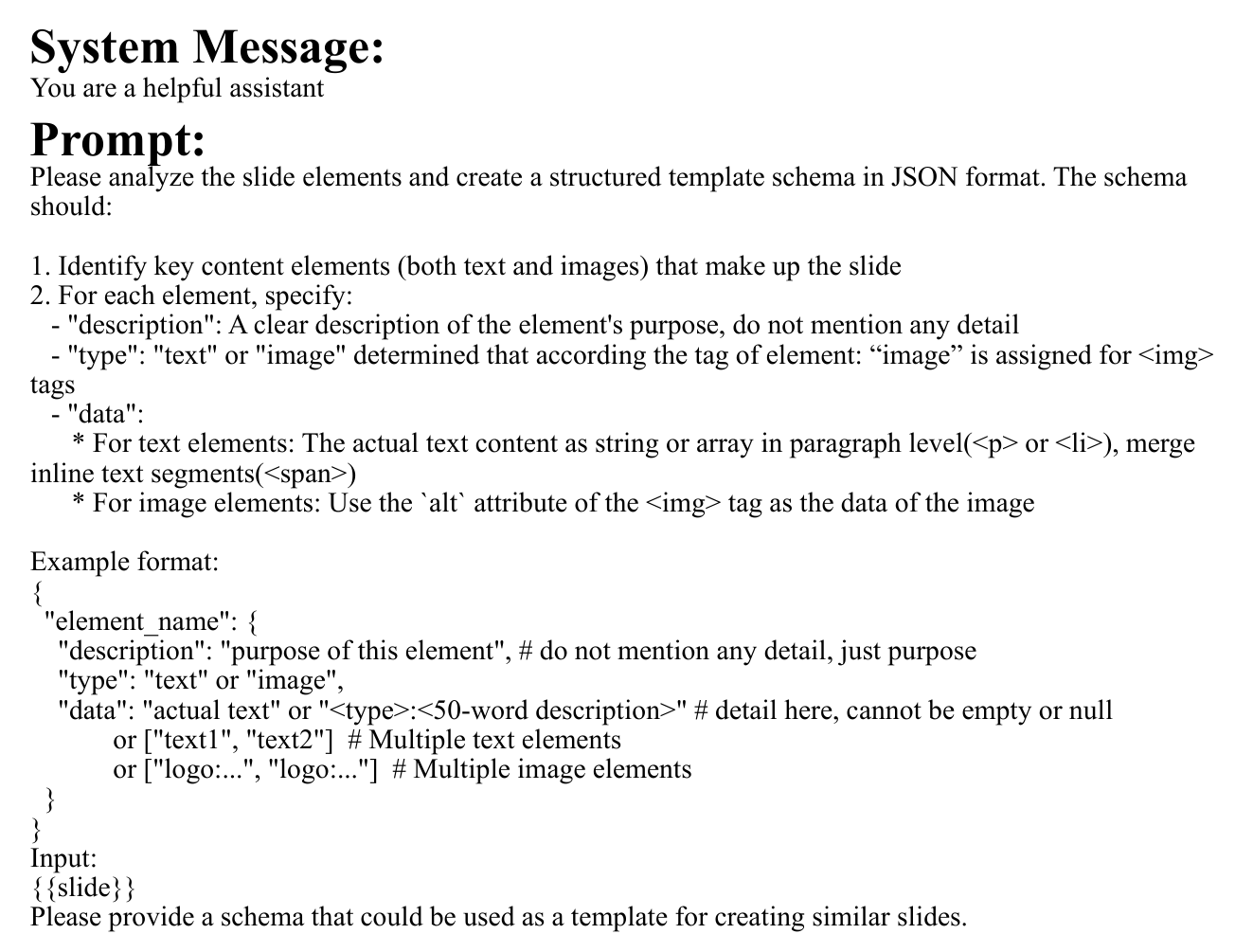}
    \caption{Illustration of the prompt used to extract the slide schema.}
    \label{fig:schema}
\end{figure}

The prompts used for presentation analysis are illustrated in Figures
\ref{fig:structure_cluster}, \ref{fig:layout_analysis}, and \ref{fig:schema}.

\subsection{Prompts for Presentation Generation}
\begin{figure}
    \centering
    \includegraphics[width=1.0\linewidth]{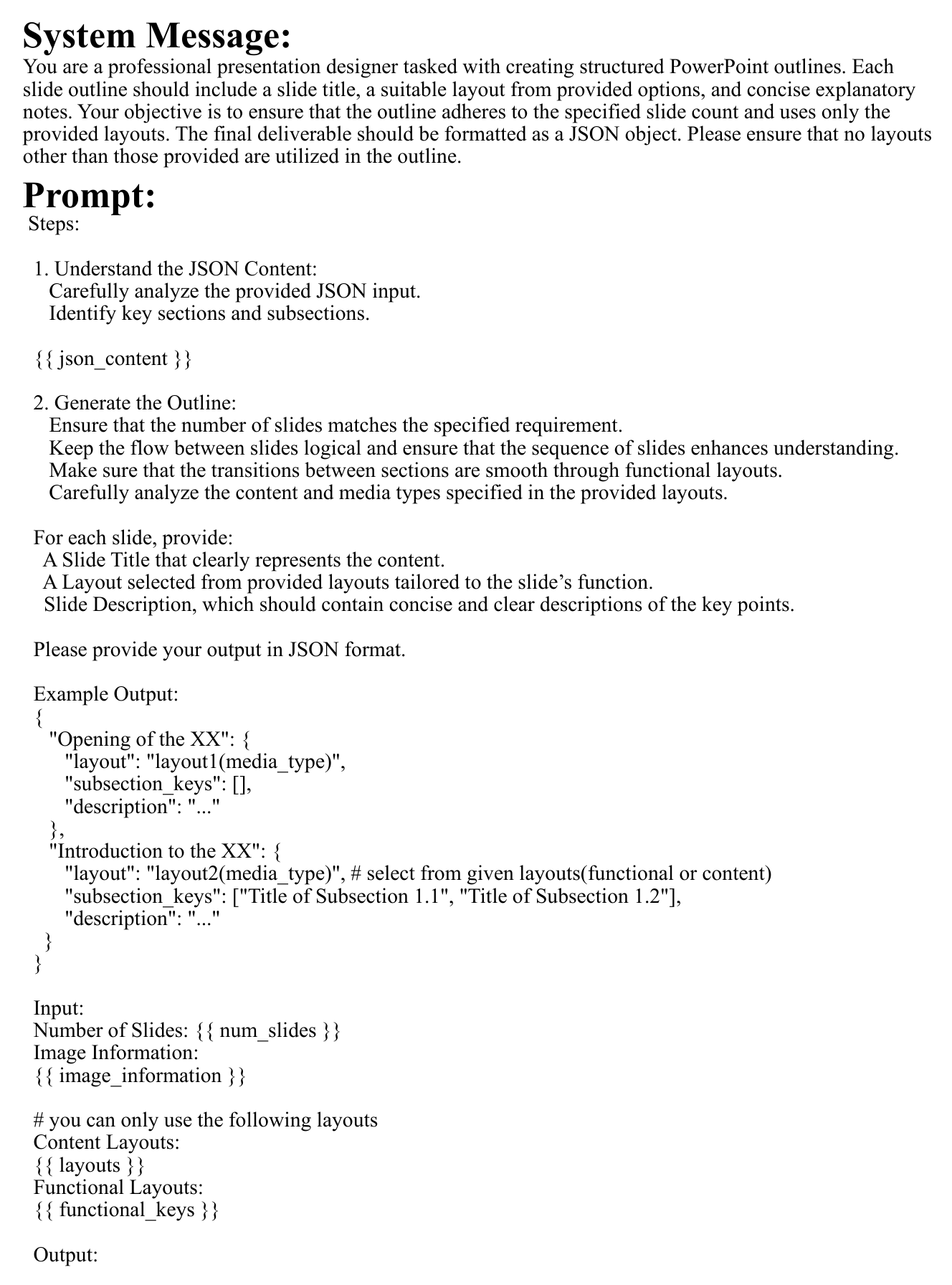}
    \caption{Illustration of the prompt used for generating the outline.}
    \label{fig:outline}
\end{figure}

\begin{figure}
    \centering
    \includegraphics[width=1.0\linewidth]{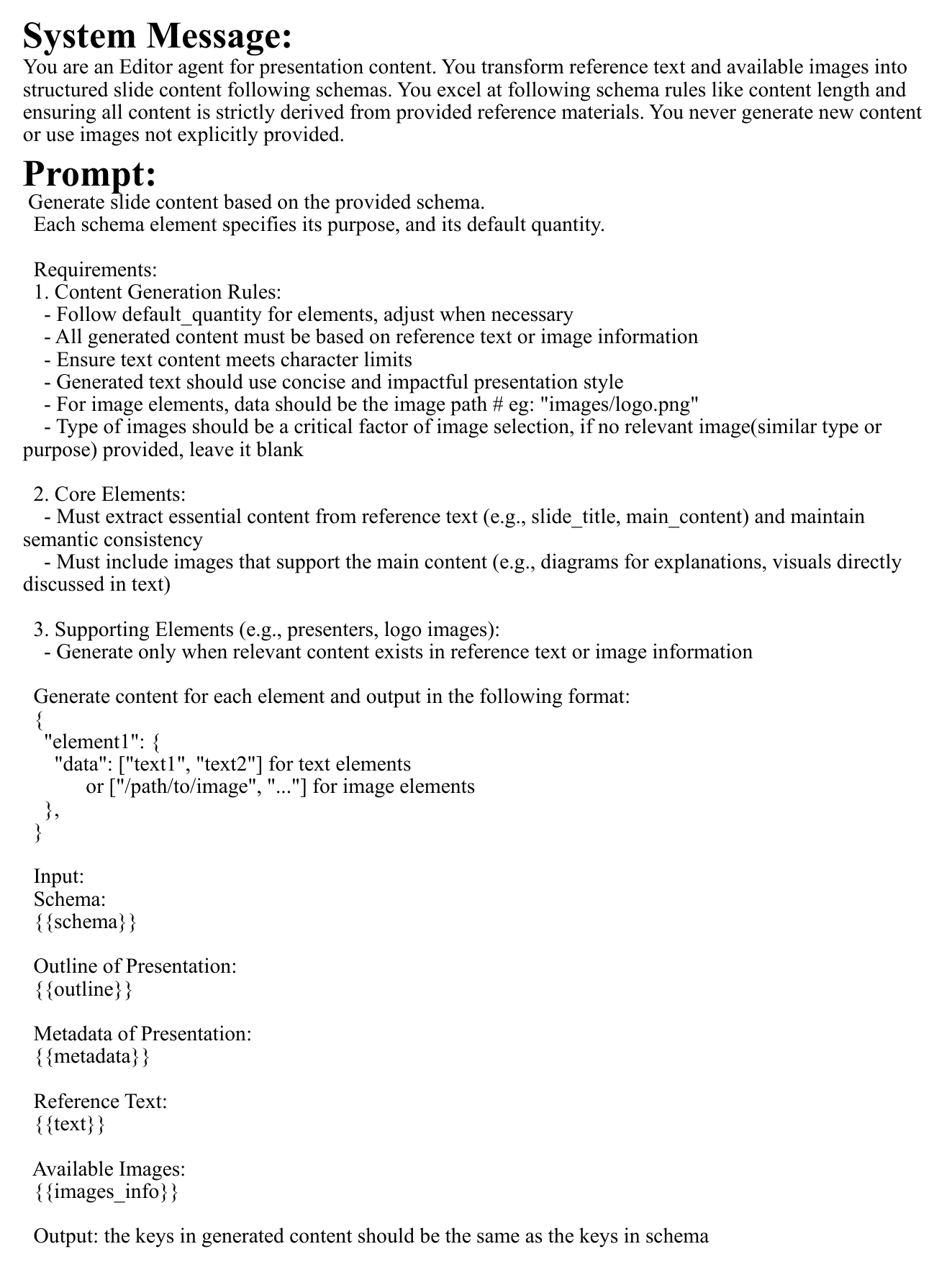}
    \caption{Illustration of the prompt used for generating slide content.}
    \label{fig:content}
\end{figure}

\begin{figure}
    \centering
    \includegraphics[width=1.0\linewidth]{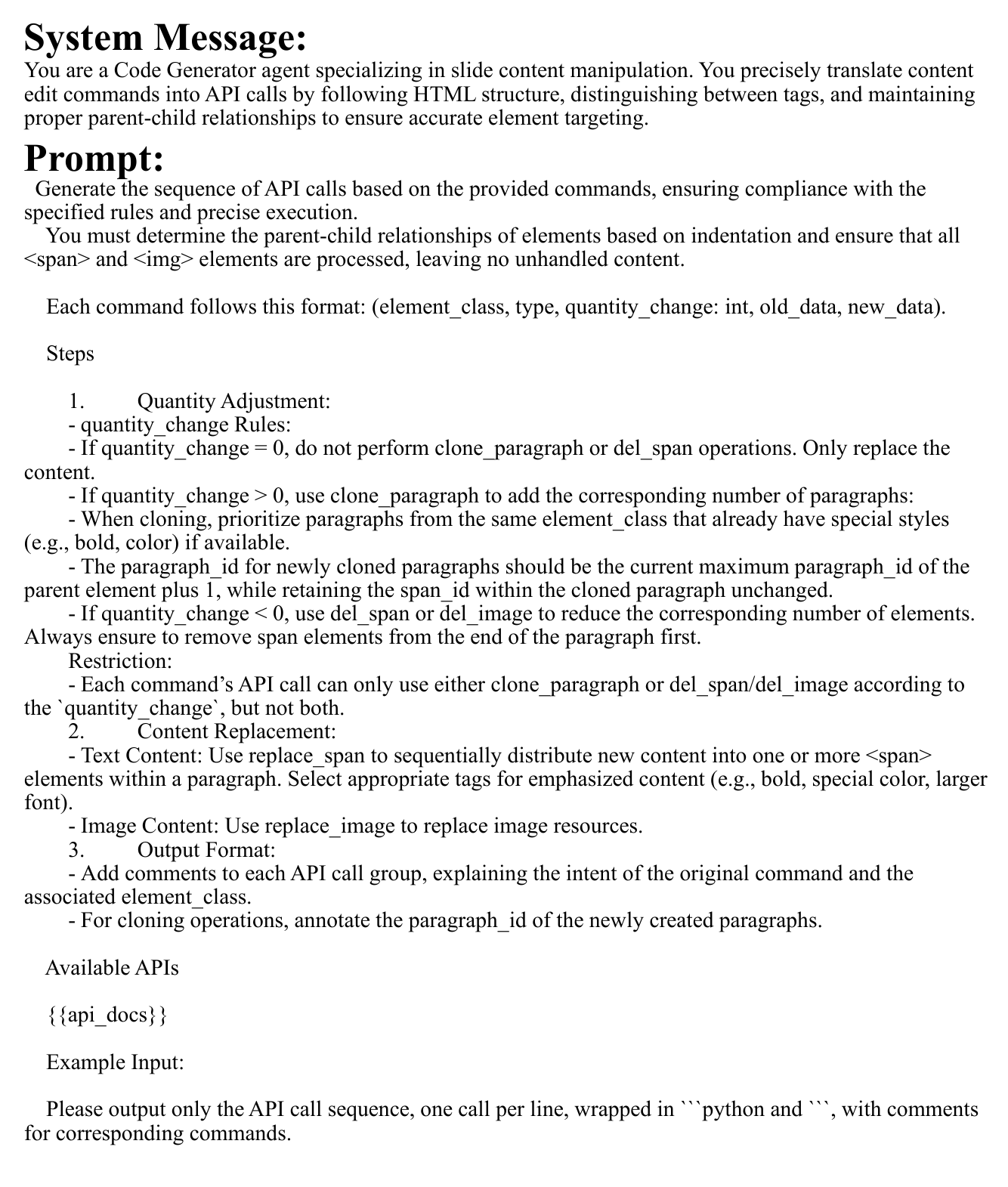}
    \caption{Illustration of the prompt used for generating editing actions.}
    \label{fig:edit}
\end{figure}

The prompts used for generating presentations are shown in Figures
\ref{fig:outline}, \ref{fig:content}, and \ref{fig:edit}.

\subsection{Prompts for \ppteval}
\begin{figure}
    \centering
    \includegraphics[width=1.0\linewidth]{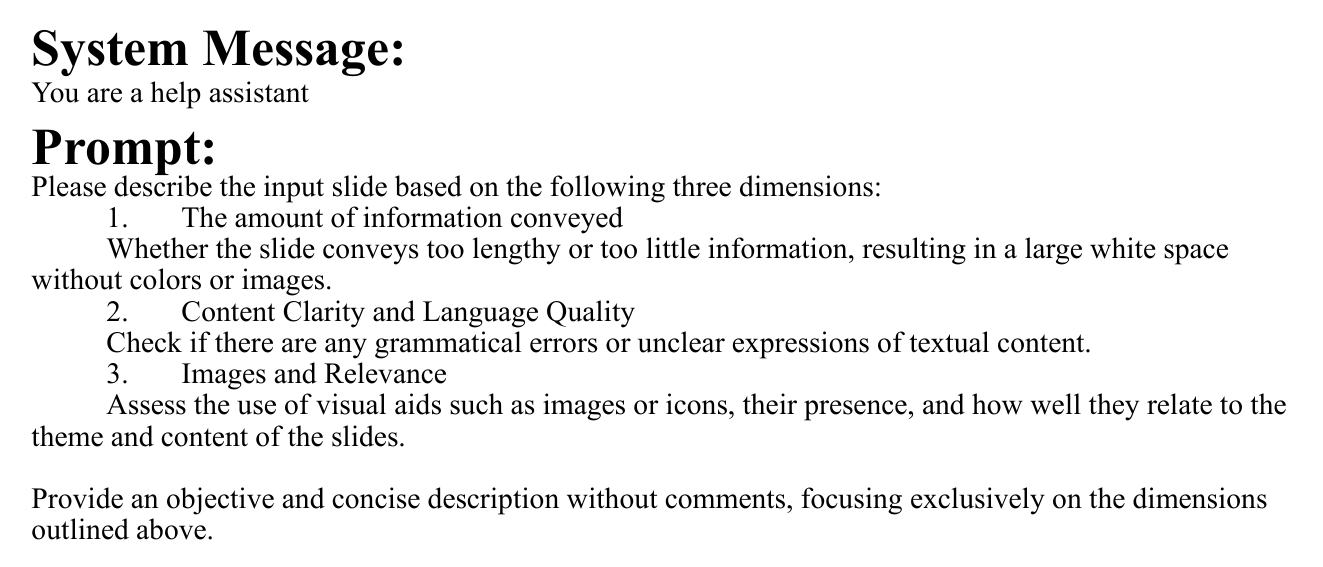}
    \caption{Illustration of the prompt used to describe content in PPTEval.}
    \label{fig:desc_content}
\end{figure}

\begin{figure}
    \centering
    \includegraphics[width=1.0\linewidth]{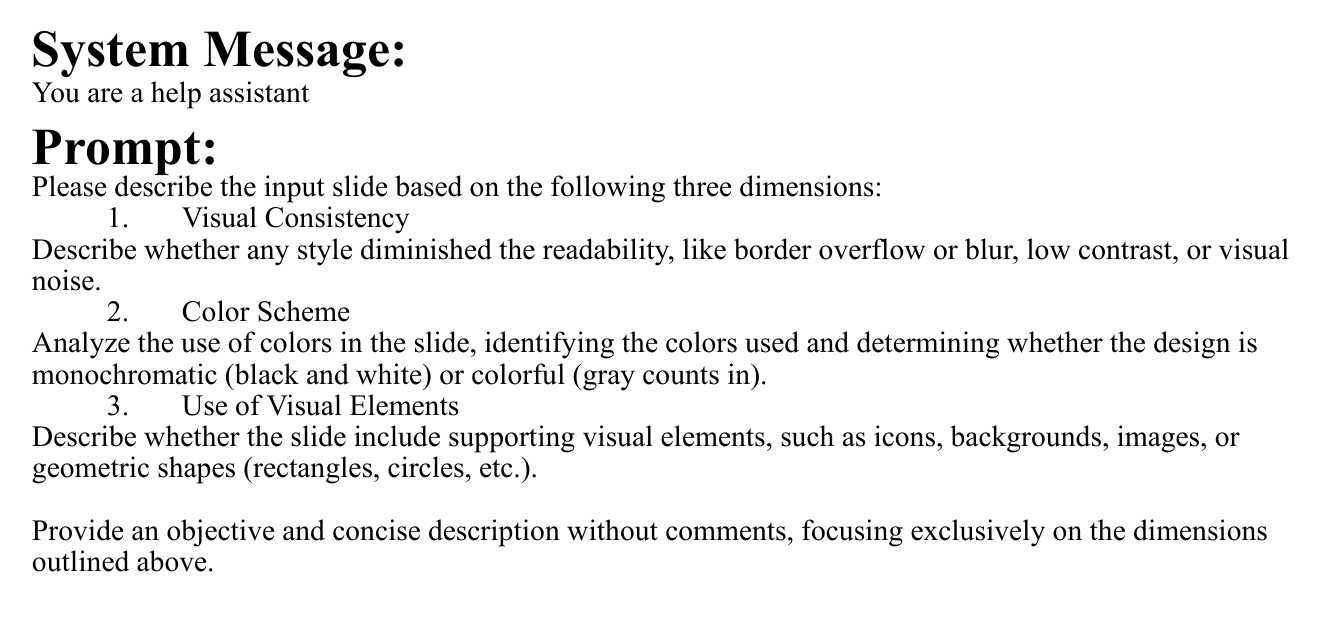}
    \caption{Illustration of the prompt used to describe style in PPTEval.}
    \label{fig:desc_style}
\end{figure}

\begin{figure}
    \centering
    \includegraphics[width=1.0\linewidth]{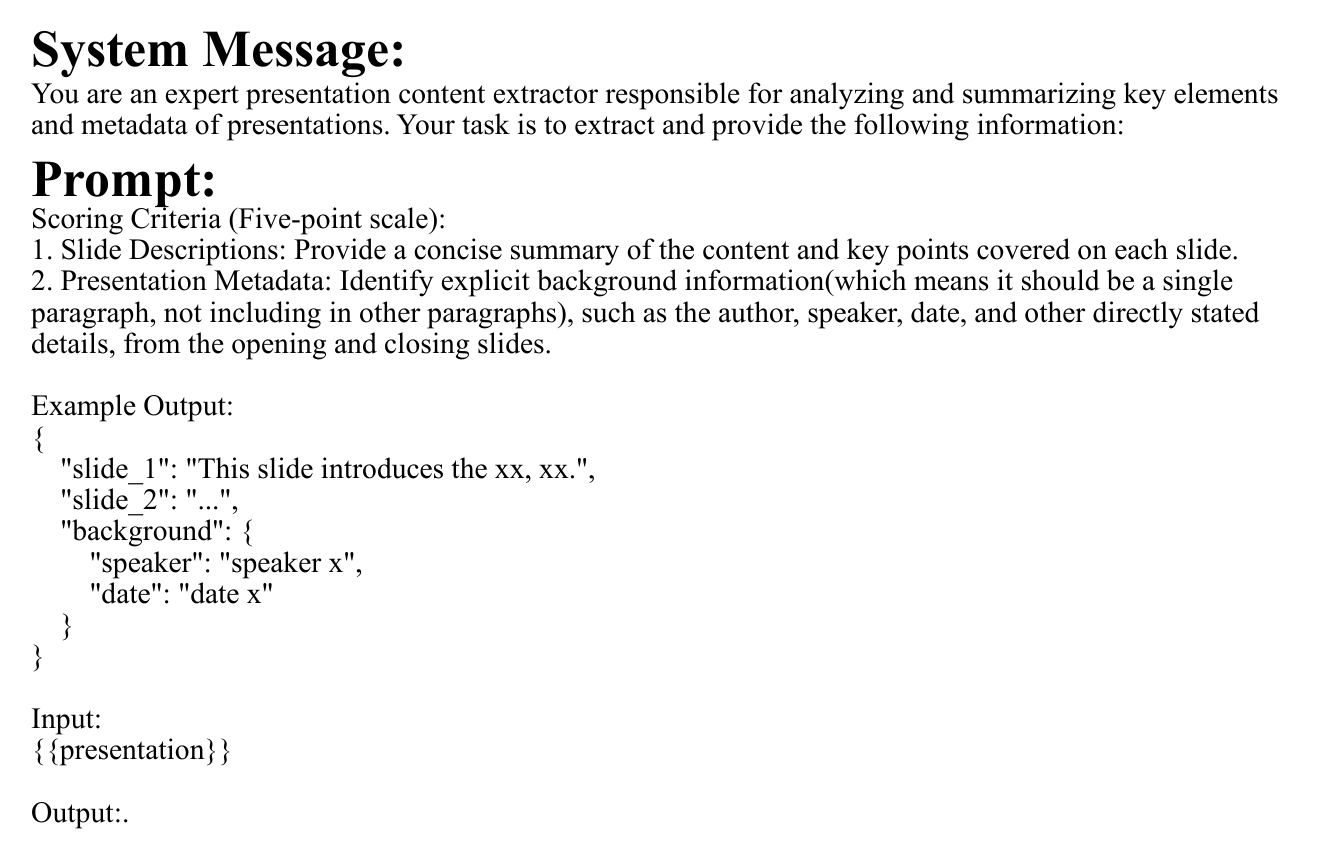}
    \caption{Illustration of the prompt used to extract content in PPTEval.}
    \label{fig:extract_content}
\end{figure}

\begin{figure}
    \centering
    \includegraphics[width=1.0\linewidth]{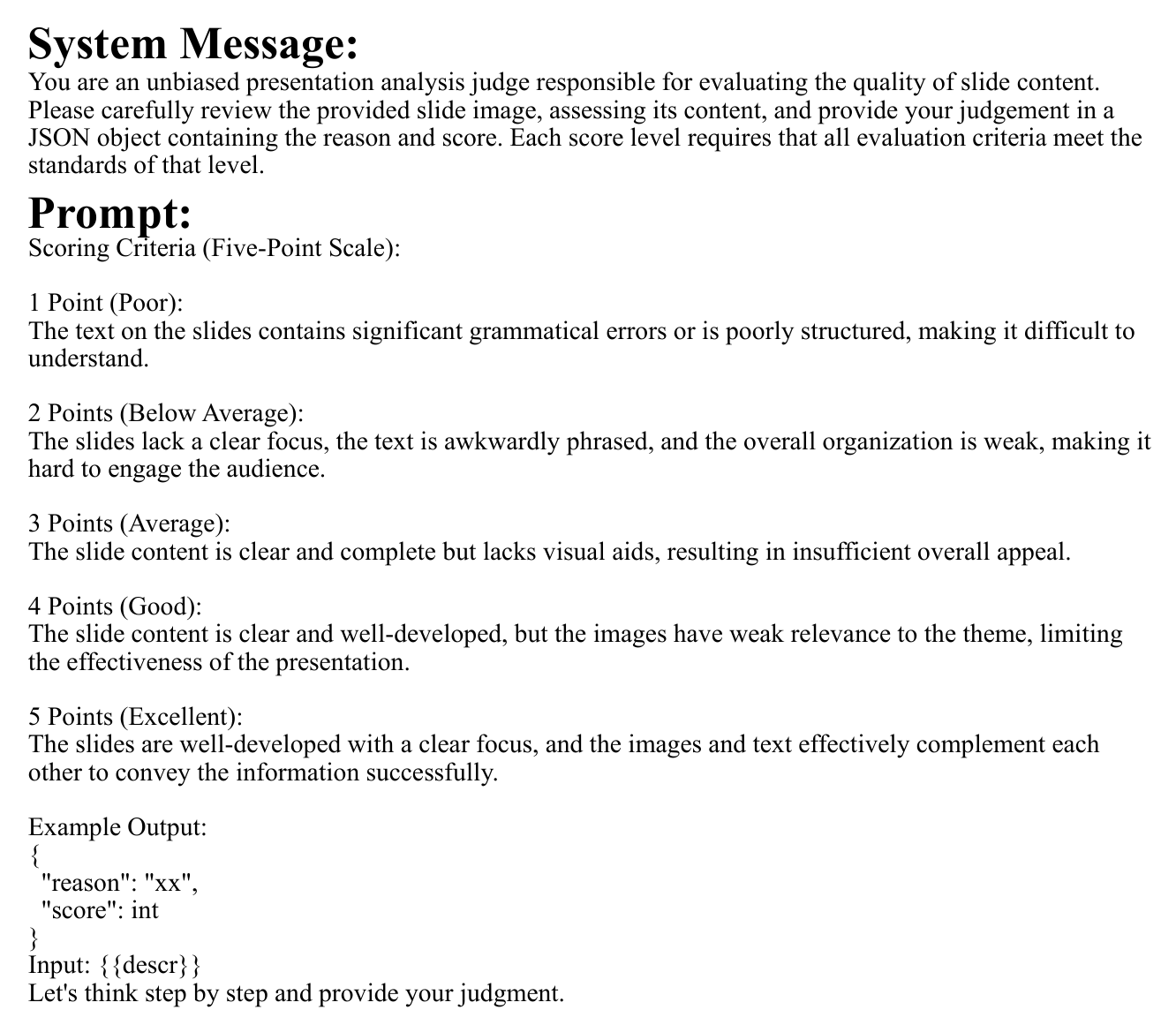}
    \caption{Illustration of the prompt used to evaluate content in PPTEval.}
    \label{fig:eval_content}
\end{figure}

\begin{figure}
    \centering
    \includegraphics[width=1.0\linewidth]{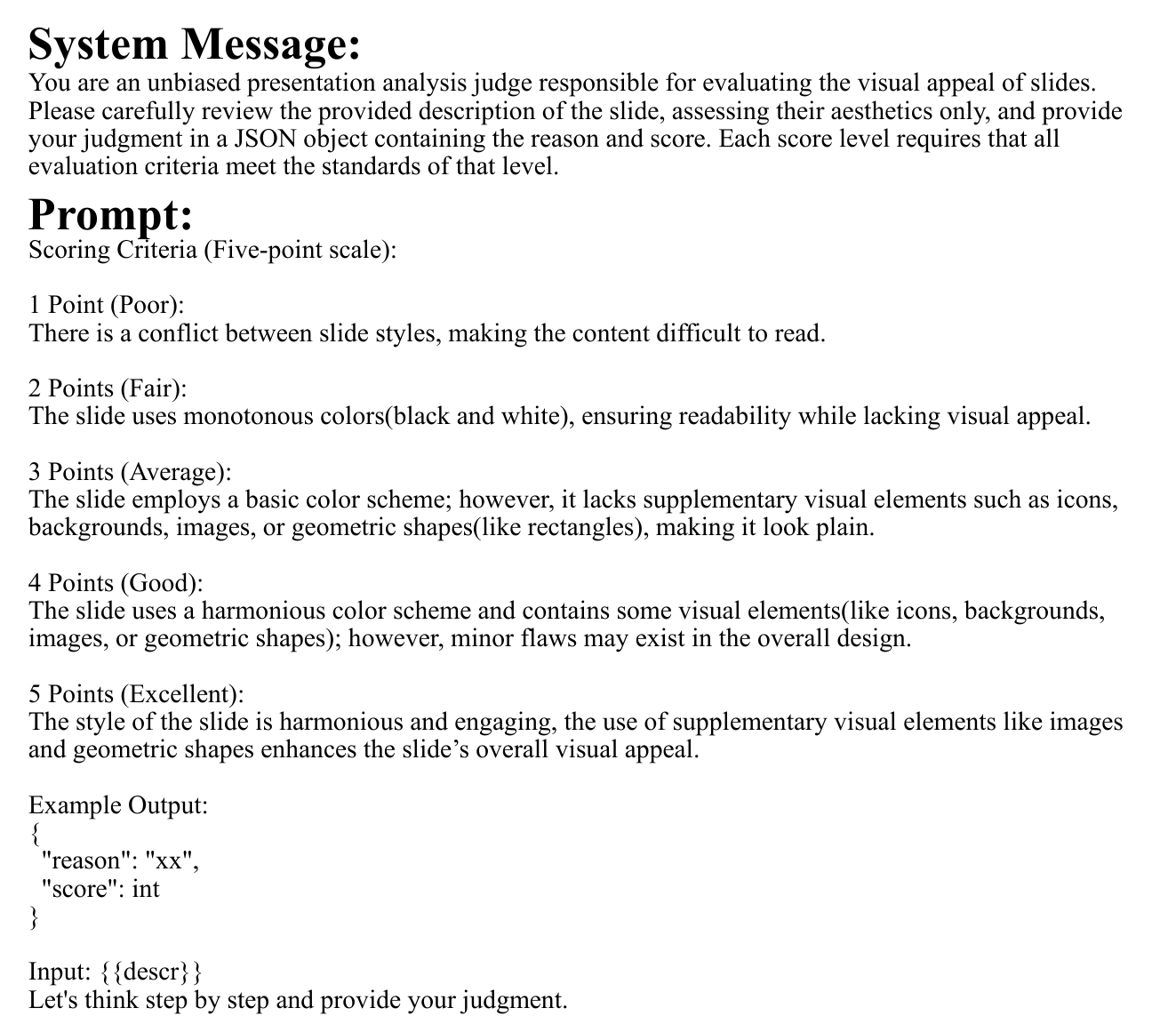}
    \caption{Illustration of the prompt used to evaluate style in PPTEval.}
    \label{fig:eval_style}
\end{figure}

\begin{figure}
    \centering
    \includegraphics[width=1.0\linewidth]{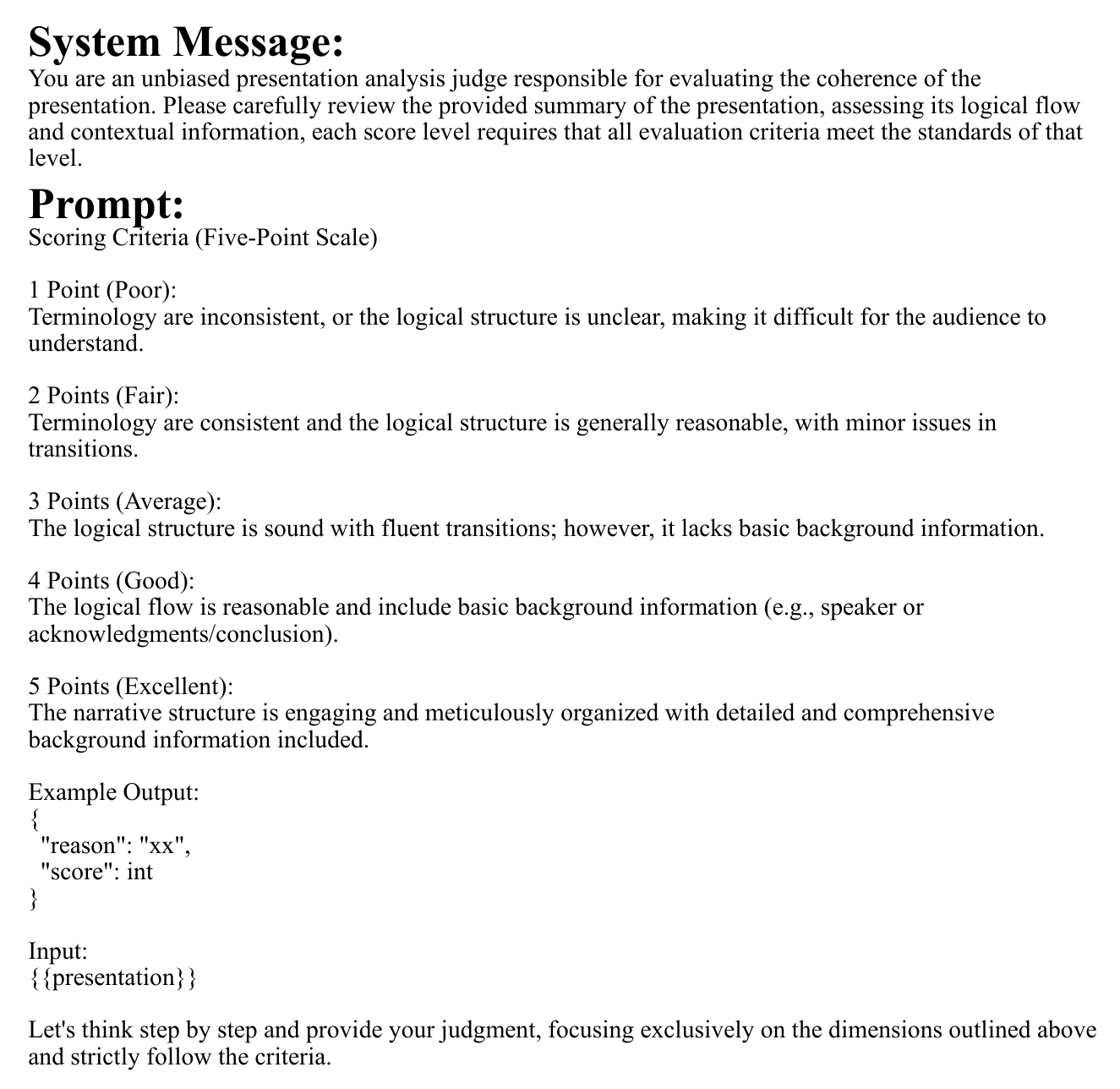}
    \caption{Illustration of the prompt used to evaluate coherence in PPTEval.}
    \label{fig:eval_coherence}
\end{figure}

The prompts used in \ppteval are shown in Figure \ref{fig:desc_content}, \ref{fig:desc_style},
\ref{fig:extract_content}, \ref{fig:eval_content}, \ref{fig:eval_style} and
\ref{fig:eval_coherence}.

\newpage

\begin{algorithm}
\caption{Slides Clustering Algorithm}
\label{alg:simplified_clustering}
\begin{algorithmic}[1]
\State \textbf{Input:} Similarity matrix of slides $S \in \mathbb{R}^{N \times N}$, similarity threshold $\theta$
\State \textbf{Initialize:} $C \gets \emptyset$

\While{$\max(S) \geq \theta$}
    \State $(i, j) \gets \arg\max(S)$ \Comment{Find the most similar slide pair}
    \If{$\exists c_k \in C \text{ such that } (i \in c_k \lor j \in c_k)$}
        \State $c_k \gets c_k \cup \{i, j\}$ \Comment{Merge into existing cluster}
    \Else
        \State $c_{\text{new}} \gets \{i, j\}$ \Comment{Create new cluster}
        \State $C \gets C \cup \{c_{\text{new}}\}$
    \EndIf
    \State \vspace{2pt} Update $S$: 
    \State \hspace{12pt} $S[:, i] \gets 0$, $S[i, :] \gets 0$
    \State \hspace{12pt} $S[:, j] \gets 0$, $S[j, :] \gets 0$
\EndWhile

\State \textbf{Return:} $C$
\end{algorithmic}
\end{algorithm}

\newpage

\begin{table*}[t]
    \centering
    
    \resizebox{\linewidth}{!}{
    \begin{tabular}{llcccccccc}
        \toprule
        \multicolumn{2}{c}{\textbf{Configuration}} & \multicolumn{4}{c}{Existing Metrics} & \multicolumn{4}{c}{PPTEval} \\
        \cmidrule(lr){3-6} \cmidrule(lr){7-10}
        Language Model & Vision Model 
        & SR(\%)$\uparrow$ & PPL$\downarrow$ &  ROUGE-L $\uparrow$  &FID$\downarrow$ 
        & \redtext{Content}$\uparrow$ & \greentext{Design}$\uparrow$ & \yellowtext{Coherence}$\uparrow$ & Avg.$\uparrow$ \\
        \midrule
        \rowcolor{gray!20} \multicolumn{10}{c}{\textit{DocPres (rule-based)}} \\
        GPT-4o$_{\texttt{LM}}$ & \multicolumn{1}{c}{--} 
        & -- & 76.42 & 13.28 & -- & 2.98 & 2.33 & 3.24 & 2.85 \\
        Qwen2.5$_{\texttt{LM}}$ & \multicolumn{1}{c}{--} 
        & -- & 100.4 & 13.09 & -- & 2.96 & 2.37 & 3.28 & 2.87 \\
        \midrule
        \rowcolor{gray!20} \multicolumn{10}{c}{\textit{KCTV (template-based)}} \\
        GPT-4o$_{\texttt{LM}}$ & \multicolumn{1}{c}{--} 
        & 80.0 & \underline{68.48} & 10.27 & -- & 1.99 & 2.35 & 2.85 & 2.40 \\
        Qwen2.5$_{\texttt{LM}}$ & \multicolumn{1}{c}{--} 
        & 88.0 & \textbf{41.41} & \textbf{16.76} & -- & 2.24 & 2.59 & 2.95 & 2.59 \\
        \midrule
        \rowcolor{gray!20} \multicolumn{10}{c}{\textit{\pptagent (ours)}} \\
        GPT-4o$_{\texttt{LM}}$ & GPT-4o$_{\texttt{VM}}$ 
        & \textbf{97.8} & 721.54 & 10.17 & 7.48 & \textbf{3.17} & 
    \textbf{3.16} & \underline{4.20} & \textbf{3.54} \\
        Qwen2-VL$_{\texttt{LM}}$ & Qwen2-VL$_{\texttt{VM}}$ 
        & 43.0 & 265.08 & 13.03 & \underline{7.32} & 1.34 & 1.43 & 1.75 & 1.50 \\
        Qwen2.5$_{\texttt{LM}}$ & Qwen2-VL$_{\texttt{VM}}$ 
        & \underline{95.0} & 496.62 & \underline{14.25} & \textbf{6.20} & \underline{3.11} & \underline{3.10} & \textbf{4.25} & \underline{3.48} \\
        \bottomrule
    \end{tabular}}
    \caption{Weighted Performance comparison of presentation generation methods, including
   DocPres, KCTV, and our proposed \pptagent. 
   Results are evaluated using Success Rate (SR), Perplexity (PPL), Rouge-L, Fr'{e}chet Inception Distance (FID), and SR-weighted PPTEval.}
    \label{table:weighted_mainexp}
\end{table*}
\begin{table}[h]
   \centering
   
\resizebox{\linewidth}{!}{
\begin{tabular}{lccccc}
       \toprule \textbf{Setting} & SR(\%) & \redtext{Content} & \greentext{Design} & \yellowtext{Coherence} & Avg. \\
       \midrule
       \pptagent & \textbf{95.0} & 3.11 & 3.10 & 4.25 & 3.48  \\
       \textsl{w/o Outline} & 91.0 & 2.94 & 3.00 & 3.05 & 3.00 \\
       \textsl{w/o Schema} & 78.8 & 2.42 & 2.54 & 3.18 & 2.71 \\
       \textsl{w/o Structure} & \underline{92.2} & \textbf{3.02} & 2.99 & 3.18 & 3.06 \\
       \textsl{w/o CodeRender} & 74.6 & 2.43 & 2.49 & 3.26 & 2.73 \\
    \bottomrule
   \end{tabular}}
\caption{Ablation analysis of \pptagent utilizing the Qwen2.5$_{\texttt{LM}}$+Qwen2-VL$_{\texttt{VM}}$ configuration, with PPTEval scores weighted by success rate to demonstrate each component's contribution.}
      \label{table:weighted_ablation}
      
   \end{table}
\begin{table}[h]
\centering
\begin{adjustbox}{max width=\linewidth} 
\begin{tabular}{lcccc}
\toprule
Domain & SR (\%)  & PPL & FID & PPTEval \\
\midrule
Culture & 93.0 &185.3&5.00& 3.70 \\ 
Education & 94.0 &249.0&7.90& 3.69 \\ 
Science & 96.0 &500.6&6.07& 3.56 \\ 
Society & 95.0 &396.8&5.32& 3.59 \\ 
Tech & 97.0 &238.7&6.72& 3.74 \\ 
\bottomrule
\end{tabular}
\end{adjustbox}
\caption{Evaluation results under the configuration of Qwen2-VL\(_{\texttt{LM}}\)+Qwen2-VL\(_{\texttt{VM}}\) in different domains, using the success rate (SR), PPL, FID and the average PPTEval score across three evaluation dimensions.}
\label{table:domain_detail}
\end{table}

\newpage


\end{document}